\documentclass[sigconf]{acmart}

\AtBeginDocument{%
  }

\usepackage{xspace}
\usepackage{enumitem}
\usepackage{microtype}
\usepackage{balance}
\usepackage[textsize=footnotesize,textwidth=15mm,backgroundcolor=black!1]{todonotes}
\usepackage{multirow}
\usepackage{color, colortbl}
\usepackage{graphicx}
\usepackage[skip=0pt]{caption}

\usepackage{float}
\usepackage{xcolor}
\usepackage{booktabs} 
\usepackage{algpseudocode}
\usepackage{graphicx}
\usepackage{subfigure}
\usepackage{adjustbox}
\usepackage{cleveref}

\usepackage{cancel}

\usepackage{pifont}

\definecolor{darkgreen}{rgb}{0.0, 0.5, 0.0}
\definecolor{babyblueeyes}{rgb}{0.63, 0.79, 0.95}
\newcommand{\cmark}{\textcolor{darkgreen}{\ding{51}}}

\newcommand{\xmark}{\textcolor{red}{\ding{55}}}

\usepackage[ruled, vlined]{algorithm2e}
\usepackage{algpseudocode}

\usepackage{lipsum}  

\usepackage[compact]{titlesec}

\newcommand{\noindpar}[1]{\noindent {\bf #1}}

\newcommand{\mypar}[1]{\noindent {\em #1}}

\newcommand{\sysname}{FreeML\xspace}

\newcommand{\compname}{SparseComp\xspace}

\newcommand{\exitname}{gNet\xspace}

\setcopyright{rightsretained} \copyrightyear{2024}
\acmISBN{}
\acmDOI{}

\acmConference[EWSN'24]{The 21st International Conference on Embedded Wireless Systems and Networks}{10th-13th December 2024}{Abu Dhabi, UAE}

\usepackage{array}
\usepackage{adjustbox}
\newcolumntype{C}[1]{>{\centering\let\newline\\\arraybackslash\hspace{0pt}}m{#1}}

\begin{document}

\title{
 Memory-efficient Energy-adaptive Inference of {Pre-Trained Models} on Batteryless Embedded Systems  
}

\settopmatter{authorsperrow=4}

\author{Pietro Farina}
\authornote{Both authors contributed equally to the paper}
\affiliation{%
 \institution{University of Trento}
 \city{Trento}
 \country{Italy}}
\email{pietro.farina@studenti.unitn.it} 
\author{Subrata Biswas}
\authornotemark[1]
\affiliation{%
 \institution{Worcester Polytechnic Institute}
 \city{Worcester}
 \country{USA}}
\email{sbiswas@wpi.edu} %
\author{Eren Y{\i}ld{\i}z}
\affiliation{%
 \institution{Ege University}
 \city{Izmir}
 \country{Turkiye}}
\email{eren.yildiz@ege.edu.tr} %
\author{Khakim Akhunov}
\affiliation{%
 \institution{University of Trento}
 \city{Trento}
 \country{Italy}}
\email{khakim.akhunov@unitn.it}

\author{Saad Ahmed}
\affiliation{%
 \institution{Georgia Institute of Technology}
 \city{Atlanta}
 \country{USA}}
\email{sahmed@gatech.edu}

\author{Bashima Islam}
\affiliation{%
 \institution{Worcester Polytechnic Institute}
 \city{Worcester}
 \country{USA}}
\email{bislam@wpi.edu}

\author{Kas{\i}m Sinan Y{\i}ld{\i}r{\i}m}
\affiliation{%
  \institution{University of Trento}
  \city{Trento}
  \country{Italy}}
\email{kasimsinan.yildirim@unitn.it}

\settopmatter{printfolios=true,printacmref=false}
\begin{abstract}

Batteryless systems frequently face power failures, requiring extra runtime buffers to maintain inference progress and leaving only a memory space for storing ultra-tiny deep neural networks (DNNs).
Besides, making these models responsive to stochastic energy harvesting dynamics during inference requires a balance between inference accuracy, latency, and energy overhead. Recent works on compression mostly focus on time and memory, but often ignore energy dynamics or significantly reduce the accuracy of pre-trained DNNs. 
Existing energy-adaptive inference works modify the architecture of pre-trained models and have significant memory overhead. Thus, energy-adaptive and accurate inference of pre-trained DNNs on batteryless devices with extreme memory constraints is more challenging than traditional microcontrollers.

We combat these issues by proposing FreeML, a framework to optimize pre-trained DNN models for memory-efficient and energy-adaptive inference on batteryless systems. FreeML comprises (1) a novel compression technique to reduce the model footprint and runtime memory requirements simultaneously, making them executable on extremely memory-constrained batteryless platforms; 
and (2) the first early exit mechanism that uses a single exit branch for all exit points to terminate inference at any time, making models energy-adaptive with minimal memory overhead.
Our experiments showed that FreeML reduces the model sizes by up to $95 \times$, supports adaptive inference with a $2.03-19.65 \times$ less memory overhead, and provides significant time and energy benefits with only a negligible accuracy drop compared to the state-of-the-art. 

\end{abstract}

\maketitle

\renewcommand{\shortauthors}{Farina et al.}

\section{Introduction}
\label{sec:intro}

Advances in electronics and energy harvesting have given rise to batteryless devices that exclusively rely on ambient energy~\cite{bakar2022protean,desai2022camaroptera}. These devices compute {\em intermittently} due to extremely scarce and transient ambient energy, creating significant challenges in hardware and software design~\cite{yildiz2023efficient}. Despite these challenges, the application space of intermittent computing is rapidly expanding~\cite{ahmed2024internet}. 

On-device intelligence has become increasingly important for batteryless edge applications since it offers more efficient, reliable, timely, and secure computing solutions~\cite{gobieski2019intelligence, zygarde2020islam}. Deep neural network (DNN) inference is feasible on batteryless sensing platforms, and it enhances the systems efficiency and throughput considerably~\cite{zygarde2020islam,montanari2020eperceptive,bakar2022adaptive}.
However, deploying a pre-trained DNN model on an extremely resource-constrained batteryless device and running it intermittently pose significant problems to tackle. 
 
\noindpar{P1: Memory Scarcity.} Batteryless platforms are extremely memory-constrained, often having kilobyte-sized nonvolatile memory, e.g., 256KB FRAM~\cite{msp430fr59xxx}, and only a few kilobytes of SRAM. FRAM is mainly used to back up and recover program data and states for power failure-resilient intermittent computation. Besides,  inference requires extra memory for backup and recovery since input and output activations of the layers need to be preserved in FRAM~\cite{gobieski2019intelligence}. Hence, only the remaining part of the FRAM (often less than half of its total size) can be used to store
the parameters of the model. While traditional compression techniques can fit a pre-trained DNN model in 256KB by preserving its precision~\cite{lin2020mcunet, han2015deep, bhattacharya2016sparsification}, they are insufficient when the memory available to store and execute the model is significantly smaller, as they result in a significant drop in accuracy~\cite{zygarde2020islam}. On the other hand, existing works on batteryless systems obtain small models in a rigid and costly way: generate several compressed networks, retrain them, and perform an expensive search to find the best network~\cite{gobieski2019intelligence}. 
Besides, they do not consider the energy dynamics and runtime memory requirements, which require support for different compression scales. Therefore, we need a solution to deploy pre-trained DNNs on batteryless systems smoothly by bridging the accuracy gap while matching the extreme memory and energy requirements.  

\noindpar{P2: Energy Dynamics, Memory, and Latency.} Unstable and sporadic availability of harvestable ambient energy can prevent executing inference intermittently with an acceptable latency. Existing works addressed this issue by adapting inference accuracy concerning the available energy by -- (1) maintaining multiple versions of the same model that offer different accuracies and latencies~\cite{lee2019neuro, bakar2022adaptive, bakar2022protean}, which is memory-inefficient for a batteryless device; or (2) constructing models with early exit branches~\cite{montanari2020eperceptive,zygarde2020islam}, enabling the early termination of inference and providing {\em anytime} output with reasonable accuracy. Early-exit introduces significant memory overhead to maintain parameters for each exit branch and needs to keep input activations of these networks to provide results at any time. Besides, all these approaches require either changing the model structure or retraining the whole model and thus do not support working with pre-trained DNN models. Therefore, a plug-and-play early exit solution that conforms to the extreme memory constraints by introducing minimal memory overhead without altering the baseline DNN model is required.

\noindpar{\indent Contributions.} To fill this gap, we introduce \sysname---a systematic pipeline to optimize pre-trained DNN models for memory-efficient and energy-adaptive inference on batteryless systems. \sysname comes with the following specific features:

\mypar{ (1) Sparsity-imposed DNN Compression  (\textbf{\compname})} is a new iterative algorithm that compresses selective layers of a pre-trained DNN by retraining and imposing sparsity constraints simultaneously. \compname minimizes accuracy degradation due to compression by retraining {\em only} the selected layers using a {\em small percent of the training data}. \compname does not employ expensive solutions, such as neural architecture search~\cite{gobieski2019intelligence,mendis2021intermittent}, that require retraining of the whole model and fine-grained search within a large configuration space. In addition, \compname employs layer separation to reduce the amount of runtime memory space needed to store layer activations during intermittent inference.

\mypar{ (2) Global Early Exit Network (\textbf{\exitname})} is the first plug-and-play early exit architecture that uses a single exit branch to exit from any layer of the network, introducing minimal memory overhead and eliminating the need to retrain or restructure the model. It is a one-architecture-fits-all network, which inserts only a single exit branch into the pre-trained model instead of one branch for each layer. This single exit branch takes the output from {\em all intermediate layers} as input using a unique {\em pooling mechanism} that handles missing intermediate outputs from the later layers. Therefore, it is possible to terminate DNN inference at any time by providing inference results without buffering previous layer outputs. Finally, \exitname is the first early-exit model that removes joint-training of exit branches and the DNN model, or alternating base DNN architecture, which makes it suitable for pre-trained networks.

Our experiments show that \compname can achieve even $95 \times $ compression rates on pre-trained DNN models, i.e., reducing their sizes from MBs into a few KBs without a significant accuracy drop. \exitname reduces memory overhead $2.03x-19.65x $ times by replacing multiple exit branches of traditional early-exit models with a single global branch. Moreover, \exitname reduces the inference time by $10.84\%-16.19\%$ with a median accuracy gain of 2\% compared to the traditional early-exit models. Thanks to \compname and \exitname, \sysname provides significant time and energy benefits during intermittent inference by terminating ultra-tiny DNN inference anytime and providing timely outputs.

We release \sysname as an open-source framework via~\cite{hidden-repository} to facilitate the automatic deployment and intermittent execution of DNN models, similar to the popular end-to-end frameworks Tensorflow-Micro and EdgeImpulse~\cite{hymel2022edge}, which do not support the microcontrollers~\cite{msp430fr59xxx, max78000} commonly used in batteryless systems. Our pipeline automatically converts the adapted and compressed DNN model into a set of portable source files. These files are linked with our DNN and intermittency control libraries that can execute the model intermittently and adaptively on batteryless devices.

\section{DNN Inference on Intermittent Power}
\label{sec:background}

Batteryless devices use energy harvesters to capture ambient energy and store it in their small capacitors. Due to capacitors' limited capacity and ambient energy variability, these devices experience frequent power failures.
Therefore, they perform computes {\em intermittently} by saving their computational state when power failure is imminent and restoring it when sufficient energy becomes available for resumption~\cite{yildirim2018ink,colin2016chain,ahmed2019efficient,bakar2021rehash,majid2020dynamic,yildiz2022immortal}. Several works have demonstrated the intermittent execution of {\em custom and manually-optimized} tiny DNN models in batteryless platforms~\cite{gobieski2019intelligence,zygarde2020islam,montanari2020eperceptive,lee2019neuro,mendis2021intermittent}. However, instead of resorting to these handcrafted DNN models, we need a systematic approach to adapt existing pre-trained models to conform to the severe memory and energy limitations of batteryless systems and to execute them intermittently.

\newcolumntype{C}[1]{>{\centering}m{#1}}
\begin{table*}
	\centering
	\footnotesize
	\caption{A comparison of prior works on DNN compression and adaptive execution in batteryless systems.
 }
	\begin{tabular}{
			>{\centering}m{0.24\textwidth}
			>{\centering}m{0.33\textwidth}
			m{0.33\textwidth}<{\centering}
			}
		\toprule		

		\textbf{Prior Works} &
        \textbf{DNN Model Compression} &
        \textbf{DNN Model Adaptation}
  		 \\
			
		\midrule
            \rowcolor{red!3}
              Rehash~\cite{bakar2021rehash}, AdaMICA~\cite{akhunov2022adamica}, Camaroptera~\cite{desai2022camaroptera}, ImmortalThreads~\cite{yildiz2022immortal}, Neuro.ZERO~\cite{lee2019neuro}, LiteTM~\cite{bakar2022adaptive},   Protean~\cite{bakar2022protean}& 
            No \xmark & No \xmark \\ 
            \hline            
            HarvNet~\cite{jeon2023harvnet} & 
             No \xmark & Memory inefficient early exit branches, separate models for each branch  \\ 
            \rowcolor{black!3}            
            SONIC \& TAILS~\cite{gobieski2019intelligence} 
            &  Compressing handcrafted small models via post-facto pruning, separation for only reducing model parameters, NAS for searching the best tiny model & No \xmark 
            \\ 
            Zygarde~\cite{zygarde2020islam} 
             &Compressing handcrafted small models via post-facto pruning and separation for reducing model parameters & Memory inefficient early exit branches, separate models for each branch \\ 
            \rowcolor{black!3}
            ePerceptive~\cite{montanari2020eperceptive} &
            Compressing handcrafted small models via post-facto pruning and separation for reducing model parameters & Memory inefficient early exit branches, separate models for each branch 
             \\        
             \hline
		\rowcolor{green!3} 
		\textbf{\sysname} (this work) & 
    Compressing {\em any pre-trained DNN models} via {\bf sparsity imposed retraining} for reducing {\em model parameters} and layer separation for reducing model {\em runtime memory requirements} $\rightarrow$ {\bf \color{red}Ultra-tiny DNN models}\cmark & {\em Plug-and-play early exit branches} via a single {\bf global exit layer} that supports anytime output without altering the baseline model $\rightarrow$ 
 {\bf \color{red}Memory-efficient early exit}\cmark 
  \\
		\bottomrule
	\end{tabular}
	\label{tab:features-comparison}
 \vspace{-1.8em}
\end{table*}

\subsection{SOTA: Deploying Pre-trained DNN Models} 

\noindent\textbf{Model Compression:}
Several studies proposed DNN compression techniques to reduce their memory footprint and computational overhead to make them run efficiently on {\em mobile systems} with MB-sized memory and power-hungry MCUs. These techniques include threshold-based pruning connections with weights~\cite{han2015deep,han2015learning} or using Fisher Information to prune unimportant connections~\cite{liu2018rethinking,lee2020fast,lee2022weight}, sparsifying fully-connected layers and separating convolutional kernels with tensor decomposition and low-rank approximation~\cite{kim2015compression, bhattacharya2016sparsification}. Some studies compress DNNs to reduce their execution time and energy consumption~\cite{yao2017deepiot,yao2018fastdeepiot}.  Unfortunately, these techniques ignore the additional memory requirement for intermittent operation, resulting in considerable accuracy drops when obtaining ultra-tiny models deployable on batteryless platforms~\cite{cai2022enable}.

Several intermittent systems~\cite{montanari2020eperceptive, zygarde2020islam, lee2019neuro} have utilized similar compression techniques on shallower and relatively smaller pre-trained DNNs by employing lower compression ratios to prevent accuracy degradation. Gobieski et al.~\cite{gobieski2019intelligence} proposed a neural architecture search (NAS)-based approach that creates several compressed configurations of a DNN (through separation and pruning) to select the most accurate and energy-efficient configuration to fit the target device. However, the NAS requires exhaustive retraining and fine-grained search within a large search space, which is time-consuming and computationally expensive. Additionally, the search space, search algorithm, and performance estimation strategy should be well-defined to obtain promising results. In summary, we
need a customizable compression technique to deploy pre-trained DNNs on batteryless systems
smoothly by bridging the accuracy gap while matching the extreme memory requirements.


\noindent\textbf{Early Exit:}
Early-exit models~\cite{teerapittayanon2016branchynet} allow dynamic reduction of inference time without sacrificing performance by introducing multiple exit branches in the network. Later works~\cite{zygarde2020islam, montanari2020eperceptive, jeon2023harvnet} expands early-exit to address the energy sporadicity of batteryless systems by trading off the accuracy, latency, and energy constraints and integrating energy as one of the deciding parameters for early-exit models. However, these works require storing multiple exit branches, increasing the memory overhead with the number of exits. Though Zygarde~\cite{zygarde2020islam} reduces the overhead by using clustering as an exit branch instead of a neural network, this still adds memory requirement for $n$ clusters and changes the network architecture to a Siamese network. HarvNet~\cite{jeon2023harvnet} uses NAS to find the optimum number of exit points, which still requires storing the parameters of multiple exit branches. Therefore, we need to consider memory as one of the constraints in designing these early-exit models.

Besides, having multiple exit branches also requires either finishing the execution of an exit branch or storing the entire output required to execute the previous exit branch in a buffer if any interruption occurs. This introduces additional computation and memory overhead to provide "any-time output." Thus, a solution is needed to support any-time output without storing the entire output in buffers or always executing the previous exit branches.

Moreover, all these works (both batteryless and non-batteryless) require either solving a joint optimization to retrain the exit branch and the baseline neural network~\cite{teerapittayanon2016branchynet, montanari2020eperceptive} or changing the baseline  architecture~\cite{zygarde2020islam} which is computationally expensive and requires access to training dataset is not always accessible. Such retraining hinders us from using pre-trained models that are gaining popularity. Therefore, a "plug and play" early exit mechanism is needed to introduce dynamic inference without altering the baseline DNN.

\subsection{Unique Features of \sysname} 

Table~\ref{tab:features-comparison} compares \sysname with previous studies that demonstrated intermittent DNN inference on batteryless systems. \sysname is the first that provides a systematic pipeline to generate energy-aware adaptive and ultra-compressed accurate models from pre-trained DNNs,  facilitating the deployment and intermittent execution of these models on real batteryless hardware platforms. Prior works are limited to the efficient intermittent execution of a few {\em handcrafted and optimized} small DNNs. \sysname addresses two main concerns simultaneously: memory footprint and energy awareness. 

\sysname introduces \compname, a new approach that casts pruning into a {\em constrained optimization problem}, which aims to maximize the model accuracy while meeting the memory constraints. This is accomplished by {\em automatically but selectively retraining} large layers using only a small subset of the training data while simultaneously {\em imposing sparsity}. This strategy avoids exhaustive NAS to find the optimal model that meets the memory footprint requirements. \compname is not a trivial algorithm since it (1) imposes a different sparsity constraint for each layer to achieve minimal accuracy degradation and (2) reduces the runtime memory requirements needed to execute the model on the device. 

In traditional early exit models, multiple exit branches need to be stored. \sysname employs \exitname, the \emph{first one-architecture-fits-all early-exit network} where a single network branch works as the exit point for all or selected~\cite{jeon2023harvnet} layers in the network. It is a "plug and play" exit model that supports "any-time output" with negligible overhead by extracting and storing the significant components of all or selected layer outputs in a single input buffer. \exitname does not require altering the baseline model and is the first work that supports inserting exit layers on any pre-trained network. However, developing and training \exitname is non-trivial for three reasons: (1) different layers’ outputs vary in shape, making a one-fit-all model harder to achieve; (2) the output of later layers is not present when exiting from a previous layer; and (3) the memory, time, and computation overhead of the exit branch needs to be minimal. 
\section{\sysname for DNN Intermittent Inference}
\label{sec:algo}

The \sysname pipeline follows a two-phase workflow. In the first phase, a given pre-trained DNN model is compressed to an ultra-tiny model ({\em \compname}) to fit into the target device's memory. Then, a global early exit network is augmented to the ultra-tiny model ({\em \exitname}) to create an energy-aware and adaptive model. The second phase involves converting the compressed adaptive DNN model into C code that is linked with the \sysname ML library, which facilitates power-failure-resilient inference on the target platform. 

\subsection{Sparsity-imposed Compression of Models}

\label{sec:compression}

\textbf{\compname} proposes a unique approach by treating DNN model compression as a {\em constrained optimization} problem. The goal of this optimization is to reduce the model's memory footprint so that it can fit into the limited memory of the target device while minimizing the accuracy drop that occurs due to compression. To solve this optimization problem, \compname selectively {\em retrains} the large layers of the pre-trained DNN model while imposing {\em autonomously identified} sparsity constraints during training. In addition to reducing the number of parameters, \compname minimizes the model's runtime memory requirements during intermittent inference by reducing the memory needed to store intermediate results. 

\subsubsection{Overview.} \compname includes two main components:

\par {\em (1) Runtime-Aware Separation of Layers}: To enable failure-atomic intermittent execution of DNN inference, \sysname maintains an internal {\em working buffer} in nonvolatile memory whose size is sufficient to keep the corresponding layer's both input and output activations. This is mandatory to make inference computations idempotent so that \sysname never reads and then writes to the same memory location, eliminating memory inconsistencies due to write-after-read (WAR) dependencies~\cite{yildiz2022immortal}. \compname applies {\bf separation} techniques~\cite{gobieski2019intelligence,bhattacharya2016sparsification} to split the layers that increase working buffer requirements, decreasing their runtime memory overhead with a minimal drop in model accuracy.  

\begin{figure}
    \centering
    \includegraphics[width=\columnwidth]{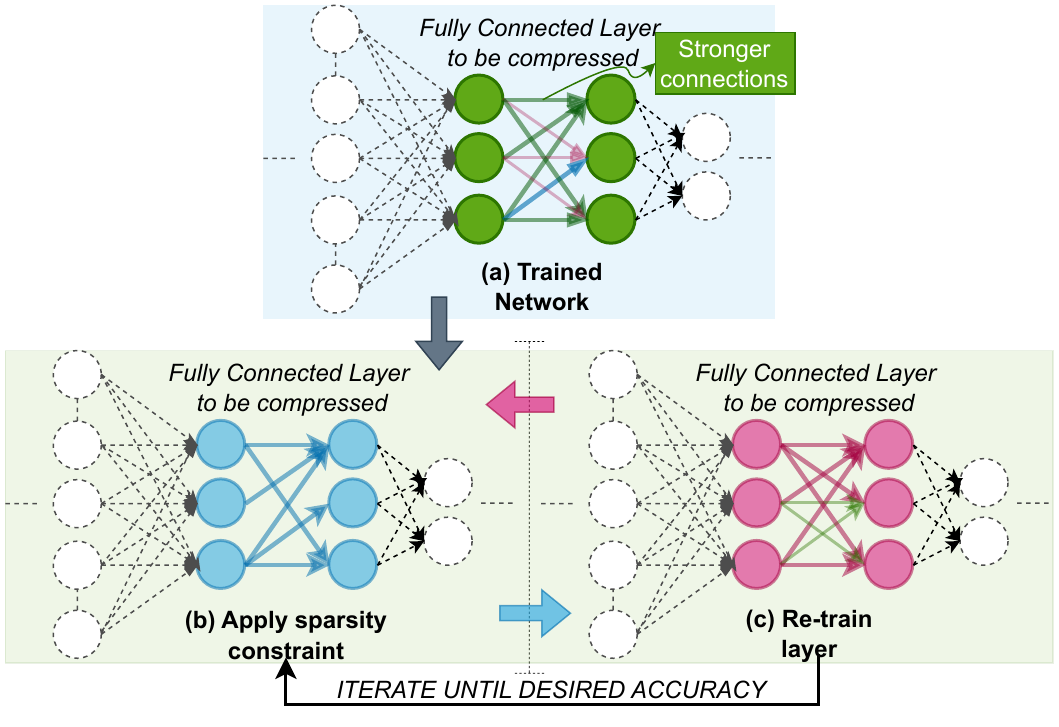}
    \caption{The \compname compression scheme. Smaller weights are pruned after imposing the sparsity constraint. The pruned weights can appear in the next epoch again during re-training. After several iterations, the layer is compressed with a minimal drop in the model accuracy.}
    \label{fig:compression-flow}
    \vspace{-1em}
\end{figure}

{\em (2) Iterative Unstructured Pruning}: After separating the layers of the pre-trained DNN model and reducing its runtime memory requirement, the next step is systematic pruning to compress and fit that model into the target device's memory. To achieve this, \compname follows a process where it selects the largest fully connected or convolutional layer in the model and defines a {\bf sparsity constraint} for it. To compress the selected layer, we freeze all the preceding layers, i.e., their weights, biases, or filters are not modified, and only retrain the selected layer and the following ones by imposing the selected sparsity constraint {\bf only} on the layer to be compressed. We impose the sparsity constraint using {\em projected gradient descent} that employs iterative hard-thresholding~\cite{gupta2017protonn}, allowing the memory constraints to always be respected on the selected layer during retraining. Once the selected large layer is retrained to achieve the desired sparsity with minimal degradation in the model accuracy, we start a new re-training iteration and repeat the process until the desired model size requirements are met.

\subsubsection{Runtime-Aware Separation of Layers} During deep neural network (DNN) inference at runtime, a common step of executing these layers is the multiply-and-accumulate (MAC) operation, which involves computing the product of two numbers and adding that product to an accumulator ($x+=y*z$). However, performing MAC operations can cause anti-dependencies (i.e., WAR dependencies)~\cite{gobieski2019intelligence,colin2016chain} since they need to read and write to the same memory locations in non-volatile memory, i.e., the accumulator \texttt{x}. If these operations are repeated due to power-failure interruptions, they can lead to different results due to anti-dependencies, making them inherently non-idempotent (i.e., not power-failure-resilient). 

To address this issue, the \sysname library has a dedicated working buffer in non-volatile memory to maintain input and output activations of layers separately during inference at runtime. This separation ensures that \sysname never reads and then writes to the same memory locations, thereby enabling power-failure-resilient execution of each layer. The runtime memory requirement of a layer is the sum of its input and output activations. Therefore, the layer with the maximum runtime buffer size requirement specifies the size of the \sysname working buffer. \sysname employs the {\em separation} of layers to decrease their input/output sizes and, in turn, to reduce the working buffer size.  

To reduce the working buffer requirements, \compname uses separation techniques to separate convolution layers with the Tucker tensor decomposition and fully connected layers with singular value decomposition~\cite{bhattacharya2016sparsification,gobieski2019intelligence}. Previous approaches have mainly focused on reducing the number of multiplications and improving computational efficiency, but our approach also considers working buffer sizes. \compname starts with the layer that requires the largest buffer and separates it. For instance, a fully connected layer with dimensions  $m \times n$ is factorized into two layers with dimensions $m \times k$ and $k \times n$ layers where $k<m$.  In the prior case, the working buffer requirement for the layer is $m+n$, while in the latter, it is $m+k$ if we assume $m>n$. Separating layers reduces the working buffer size and energy requirements, but this comes at the cost of a drop in accuracy. We used Bayesian matrix factorization~\cite{nakajima2011global} to estimate and select the dimension of the inserted layer (i.e., $k$).

\subsubsection{Iterative Unstructured Pruning}

After reducing the working buffer requirements of the model through the separation of the layers, \compname starts compressing the pre-trained DNN model layer by layer. Instead of pruning all layers non-selectively, \compname applies a specific compression rate to each layer and removes the risk of over-pruning the most meaningful layers. The compression of a layer is a two-step iterative process: First, a sparsity ratio is {\em automatically} selected for the layer to be compressed, and then the model accuracy is {\em optimized} by retraining that layer and subsequent layers in the model by {\em continuously imposing} the selected layer sparsity. These steps are repeated until the best sparsity ratio that does not significantly degrade model accuracy is selected to compress the layer. \compname then proceeds to the next layer that contributes the most to memory requirements and employs the same steps until the desired compression rate is achieved. Throughout the compression process, only one layer changes in size with each compression iteration. This approach allows \compname to assess the impact of compression on accuracy and adjust the compression rate for layers that exhibit significant accuracy drops.

\noindpar{(1) Automatic Selection of Sparsity.} When choosing the sparsity value for compression, i.e., the number of zero-valued elements divided by the total number of elements, \compname considers the size of the layer to be compressed, the overall model size, and the target size of the compressed model. To prune larger layers, which contribute more to the model size, \compname sets the initial sparsity to 0.9 (the percentage of non-zero model parameters), resulting in a 90\% weight pruning. For smaller layers, \compname selects a higher initial sparsity value. After retraining the layer, \compname evaluates the model's accuracy. If the drop in accuracy is negligible, it increases the compression rate on the layer and repeats the process. When the accuracy drop exceeds a certain threshold (usually 3-5\%), \compname reduces the compression rate for the next iteration by increasing the sparsity value and moves to the next largest layer. If the accuracy drops significantly, \compname defines the layer as fragile and excludes it from future compressions. As the desired compression ratio is approached, \compname selects less significant sparsity values to avoid over-pruning.

\vspace{-1.2em}
\begin{algorithm}
  \DontPrintSemicolon
  \SetKwFunction{Forward}{forward}
  \SetKwFunction{HT}{hardThreshold}
  \small
  {\bf Inputs:}  $\mathcal{M}$---pre-trained DNN model, $\mathcal{W}$---weight matrix of the FC layer, $\mathcal{S}$---sparsity value, $\mathcal{D}$--- dataset of (X=value, Y=label) pairs. 
  
  {\bf Parameters:} $e$---SGD epochs, $B$---batch size, $\eta$---learning rate.
  
  \nl \Repeat{e epochs}{
    \nl pick $B$ samples randomly from dataset $D$

    \nl $\hat{Y} = \Forward(\mathcal{M},\mathcal{B})$ \Comment{forward pass to get predictions}
    
    
    \nl $\mathcal{W} \gets  \mathcal{W} - \eta \left( \sum_{i =0}^{B} {\nabla_{\mathcal{W}} \mathcal{L}_i(y_i, \hat{y}_i)} \right)$ \Comment{compute gradient}
        
    \nl $\mathcal{W} \gets \HT(\mathcal{W},\mathcal{S})$ \Comment{apply sparsity constraint}
    
    }    
  \caption{{\sc Compression of a Fully Connected Layer}}
  \label{algo:fc-compression}
  
\end{algorithm}
\vspace{-1.2em}

\noindpar{(2) Constrained Optimization with Re-training.} The formulation of the problem is presented in \cref{eq:sparsecomp}. Given input $x$ and the set $\mathbf{W}$ of the layers' weights, the main objective is to minimize the empirical error $\mathcal{E}_{emp}$ of the model prediction $f(x, \mathbf{W})$ where  $\| \mathcal{W} \|_0$ is the number of zero entries in $\mathcal{W} \in \mathbf{W}$ and $s_{\mathcal{W}}$ is the sparsity constraint associated to that layer.
\vspace{-1em}
\begin{equation} \label{eq:sparsecomp}
    \min_{ \forall \mathcal{W} \in \mathbf{W} : \frac{\| \mathcal{W} \|_0}{\| \mathcal{W} \|} \geq s_{\mathcal{W}}}  \mathcal{E}_{emp}(f(x, \mathbf{W})).
\end{equation}
For simplicity, we presented Algorithm~\ref{algo:fc-compression} that optimizes the sparsity of a given fully connected (FC) layer while maintaining the accuracy of the model. The weight matrix of an FC layer with $n$ input nodes and $m$ output nodes is stored as a matrix $\mathcal{W}$ of shape $n \times m$. The algorithm takes as input a pre-trained DNN model, the weight matrix of the FC layer to be compressed, the required sparsity constraint, and a small part of the training dataset of the original pre-trained DNN. Additionally, it has standard hyper-parameters such as the number of epochs, batch size, and learning rate. To reduce memory requirements, we can store the weight matrix as sparse by setting weights to zero conforming to the sparsity constraint. The algorithm follows the standard training procedure by selecting a batch of samples from the given data set and employing the forward pass (Lines 2-3). Then, it applies the projected gradient descent (lines 4-5). Firstly, the gradients with respect to the weights of the FC layer are calculated, and the weight matrix is updated by moving in the direction of the negative gradient (line 4). Then, the weight matrix is projected onto the feasible set, i.e., the sparsity constraint is imposed on the FC layer via {\em hard thresholding} procedure. Note that, to minimize accuracy degredation, the layers following the FC layer are also re-trained jointly (not presented in Algorithm~\ref{algo:fc-compression}). For these layers, the compression is employed by considering their pre-recorded sparsity values, if any. Retraining the layers of pre-trained DNNs can often result in overfitting. To address this issue, \compname introduces a regularization term and freezes the layers preceding the currently compressed layer. 


\noindpar{\em Hard thresholding procedure.} To impose a sparsity constraint, two thresholds, an upper and lower limit, are calculated. The weights between these thresholds---ones with smaller absolute values---are set to zero. As shown in \cref{fig:compression-flow}, which summarizes the compression process for a FC layer, weaker weights are pruned to satisfy the sparsity constraint. However, the pruned weights may reappear in the next epoch of the re-training procedure. With iterative pruning and retraining, the weights become more stable over time. \compname employs an identical procedure to compress the filters and biases of convolutional layers. The main idea is imposing sparsity via hard thresholding, allowing \compname to apply a single compression method regardless of layer types in the model. 

\subsection{Global Early Exit for Pre-Trained Networks}

\label{sec:gnet}

\textbf{\exitname} is a single network architecture that replaces multiple exit branches of the traditional early-exit architectures with a single global exit layer that supports anytime output from any layer of any pre-trained model. It introduces minimal memory overhead for early exit, which is crucial since memory is extremely scarce in batteryless embedded systems. This architecture not only significantly reduces the memory overhead of storing the parameters of the exit branches but also eliminates the need to store previous exit points output separately, which further reduces memory footprint. 

\subsubsection{Overview of \exitname}
\label{gnet_}
Instead of inserting one exit branch after each intermediate layer, \exitname takes the output from all intermediate layers as input. Developing such a ubiquitous exit architecture comes with $2$ unique challenges. (1) When exiting from an earlier exit point of the baseline DNN, the outputs of the following exit points are missing. Unlike traditional exit branches, where individual exit branches process the output from each exit point, \exitname requires a single fixed dimension input to the global exit branch. Thus, there is a need to compensate for the missing information from the future exit points. (2) The dimension of the output from each layer of the DNN varies, making it challenging to concatenate them directly. Besides, simply flattening all layers and then concatenating them is insufficient as it may create a bias towards the earlier layers where the output dimensions are usually larger. Moreover, a simple concatenated vector will be extremely large to store and process and will require a large fully connected layer for classification, defeating the purpose of early exit. 

\begin{figure}
    \centering
    \includegraphics[width=0.8\columnwidth]{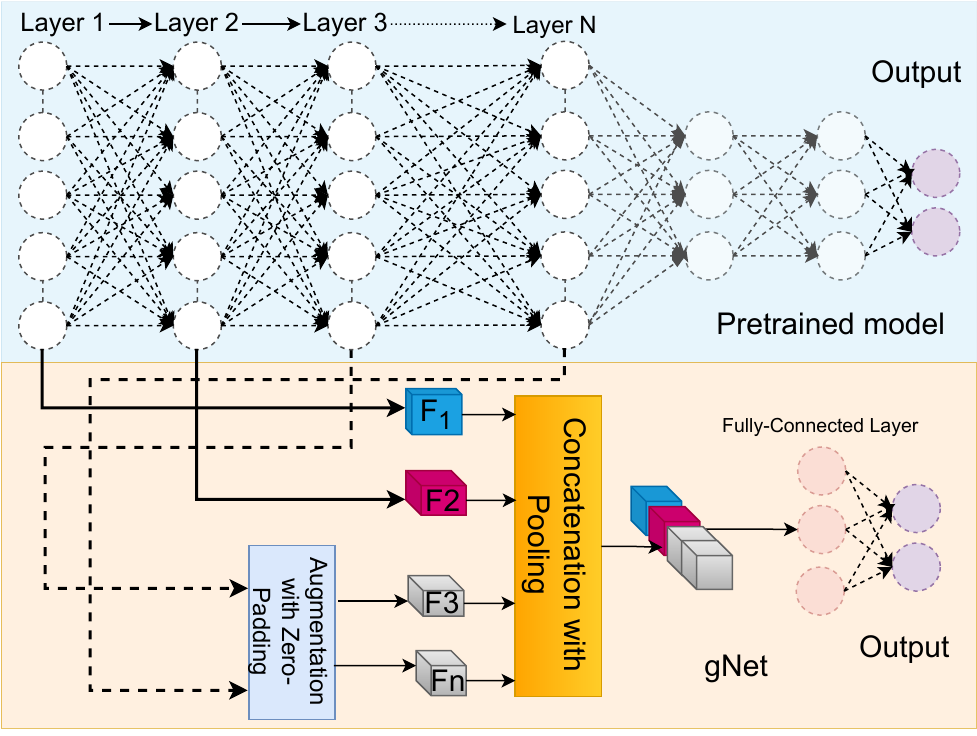}
    \caption{General overview of \exitname. In this example, early exit occurs at layer $2$; hence $F_1$ and $F_2$ are available, and the later features are zero-padded.}
    \label{fig: gnet_overview}
\vspace{-2em}    
\end{figure}

Figure \ref{fig: gnet_overview} shows the general overview of \exitname, which addresses these challenges with three major components -- (1) augmentation with zero-padding, (2) concatenation with pooling, and (3) classification with a linear layer. First, the augmentation with zero-padding handles the missing output from the layers yet to be inferred by padding them with zero to compensate for dynamic input length to the exit branch. Next, the concatenation with pooling resizes all intermediate output layers to a predefined size and concatenates them across the channel dimension. It also shrinks features across channel dimensions and extracts meaningful features when zero-padded features exist. Finally, during classification with a linear layer, the concatenated data passes through a fully connected layer and estimates which class the input belongs to. We describe the details of each of these components in the following sections.

\subsubsection{Augmentation with Zero-Padding}
The first challenge is that the output of later intermediate layers is not present while exiting from a previous layer. To compensate, we introduce the \textbf{Augmentation with Zero-Padding} layer, which performs zero-padding on the missing information from the future layers of the baseline DNN. Suppose the pre-trained network has $n$ convolution layers and only exits from the convolution layer. Thus, there are $n$ exits that are possible in maximum.  When exiting from the $i^{th}$ layer, where $i<=n$, then for $0$ to $i^{th}$ layer, we have valid intermediate outputs, and from $(i+1)^{th}$ to $n^{th}$ layer, we add zeros of the required shape. The following equation represents the resultant feature vector:
\begin{equation}
\setlength{\abovedisplayskip}{0pt}%
\setlength{\belowdisplayskip}{0pt}%
\setlength{\abovedisplayshortskip}{0pt}%
\setlength{\belowdisplayshortskip}{0pt}%
    FS = [\mathbf{F_1},\mathbf{F_2}, . . ., \mathbf{F_i}, \mathbf{0}, . . ., \mathbf{0}]
\end{equation}
Here $FS$ is the feature set, $\mathbf{F_i}$ represents intermediate output from $i^{th}$ layer. The shape of $\mathbf{F_i} $ is $C_i \times H_i \times W_i$ where $C_i$, $H_i$, and $W_i$ stand for the number of channels, height, and width of the feature.

\subsubsection{Concatenation with Pooling}
The second challenge is that the number of layers and the shape of the output tensor from each intermediate layer vary for different networks. Thus, developing a one-architecture-fit-all model is challenging. To address this, we accumulate the output of all intermediate layers to a fixed predefined shape using pooling (shown in Figure~\ref{fig: gnet_overview}). This predefined shape is chosen based on the size of all intermediate outputs and the available memory of the microcontroller. We use 3D Maxpooling to convert each output vector to a predefined shape because 3D pooling preserves the channel information which is crucial for extracting important information. Simply using the same pooling parameters for all the layers is insufficient as the earlier layers' output tensor size has a larger height and width than the later layer counterparts and can create a bias towards earlier layers. We made all layers' output the same shape to reduce bias towards any specific layer on gNet by using larger pooling kernels on the earlier layers and smaller kernel sizes for later layers. However, we do not pool the channel dimension of all the layers to the same shape because the information content in the later layers is richer over multiple channels than the earlier ones. Therefore, we use pooling kernels of various sizes for each layer, where the kernel sizes are determined based on the memory of the target microcontroller. Once all the intermediate features are the same size, we concatenate them along the channel axis and feed this fused feature to the \exitname.

\subsubsection{Classification with a Fully-Connected Layer}
The classification layer consists of a FC layer that maps the flattened vector coming from the previous layer into the predefined classes. As the FC layer performs multiplication and accumulation (MAC) operations before going into the activation layer, the order of the flattened feature set does not matter here. Over the training phase, this layer learns to put more emphasis on the available \textit{("real")} features and discard the zero-padded \textit{("fake")} features. This emphasis on real features is attained by assigning larger weights to the real feature sections and smaller weights to zero-padded sections.

\subsubsection{Agile Training of \exitname}
\label{training}
\exitname requires maintaining the output performance irrespective of the output availability from all the exit layers. 
To achieve this goal, we propose an agile training procedure.
The \exitname learns to adapt to different early exit scenarios through this training, which helps them generalize patterns, correlations, and characteristics in the different incoming data. We first develop a simulated training dataset reflecting the effect of an early exit. If a network has $n$ layers to exit from, the probability of exiting from the $i^{th}$ layer would be $\frac{1}{n}$. When exiting from the $i^{th}$ layer, where $i<=n$, then pad zeros from $(i+1)^{th}$ to $n^{th}$ layer. The training procedure of $gNet$ is shown in algorithm \ref{algo: gnet_training}. Thus, the internal network parameters are iteratively updated depending on the discrepancies between predicted and actual outputs during training to maximize the network's performance for all existing conditions. Since all the exit scenarios are encountered in the training period, we reach a global model capable of handling different exit scenarios.
\vspace{-1em}
\begin{algorithm} 
  \DontPrintSemicolon
  \SetKwFunction{Forward}{forward}
  \SetKwFunction{HT}{hardThreshold}
  \small
  {\bf Inputs:}  $\mathcal{M}$---pre-trained DNN model, $gNet$---generalized early exit model, $\mathcal{D}$--- dataset of (X=value, Y=label) pairs. 
  
  {\bf Parameters:} $e$---SGD epochs, $B$---batch size, $\eta$---learning rate, $\tau$---validation threshold, $\mathcal{L}_{val-best}$--best validation loss.
  
  \nl \Repeat{e epochs}{
    \nl pick $i^{th}$ exit layer randomly with probability of $\frac{1}{n}$ from  $n$ layers

    \nl $FS = \Forward(\mathcal{M},\mathcal{X})$ \Comment{get intermediate results}
    
     \nl $\hat{Y} = \Forward(gNet, FS)$ \Comment{get predictions}
    
    \nl $gNet \gets  gNet - \eta \left( \sum_{i =0}^{B} {\nabla_{gNet} \mathcal{L}_i(y_i, \hat{y}_i)} \right)$
        
    \nl  $\mathcal{L}_{val} \gets {\mathcal{L}(gNet(FS_{val}), Y_{val})}$ \\
     \nl \eIf{$ \mathcal{L}_{val}$ < $\mathcal{L}_{val-best}$}{
            save gNet\\
            $\mathcal{L}_{val-best} \gets \mathcal{L}_{val}$
            $count \gets 0$
        }{
            $count \gets count + 1$
        }
    \nl \eIf{$count$ >= $\tau$}{
                exit
            }{
                continue
            }
    }  
  \caption{{Training of gNet}}
  \label{algo: gnet_training}
\end{algorithm}
\vspace{-2em}
\subsection{Energy-Aware Intermittent Execution} 
\label{sec:runtime}

\sysname runtime executes ultra-tiny DNN models augmented with gNET layer-by-layer by incorporating a set of optimizations. 

\subsubsection{Buffer Reduction}
After executing a layer connected to an exit branch, \sysname performs an additional max-pooling before storing the output in a buffer. This step reduces the required buffer size by not saving the entire layer output and optimizes memory utilization while making the inputs of the early exit branch available at any time. However, max-pooling introduces a tolerable processing overhead for these layers., which can be further reduced by in-position max-pooling while calculating the output layer. 

\subsubsection{Compressed Computation}
Next, when gNET is executed through an exit branch, the \sysname runtime uses a smart approach to save time and energy by avoiding computations for zero-padded inputs that belong to layers not yet executed. This method efficiently skips MAC operations for zero-padded inputs, leading to a more efficient execution of early exit branches. Unlike traditional accelerators that only supports structured compressed computing, microcontrollers allow unstructured compressed computation, presenting us with this unique opportunity.

\subsubsection{Flexible Energy-Aware Execution Policies}
As a consequence, this strategy results in distinct computational costs for each early exit within a DNN model. Specifically, the exit layers linked to later branches tend to carry a heavier computational overhead compared to the earlier exit branches. However, they offer enhanced accuracy in return. This characteristic provides a unique opportunity for employing different policies. For instance, the application can decide to continue DNN execution for optimal accuracy or stop for reasonable accuracy by selecting an early exit branch that can output a result with a reasonable accuracy and timing overhead. Pre-recorded time and energy overheads of each layer can determine which early exit layer to choose. This gives developers the flexibility to balance processing demands, energy efficiency, and accuracy based on their application's requirements.

\section{Implementation}
\label{sec:implementation}

We implemented the proposed algorithms \compname and \exitname in Python using the PyTorch framework.
The adaptive and optimized model is automatically converted into C headers per layer including arrays holding the parameters of each layer. These platform-independent headers are combined with the ML libraries supported by existing intermittent computing runtimes, e.g., Alpaca~\cite{maeng2017alpaca} and ImmortalThreads~\cite{yildiz2022immortal}, to be executed intermittently.

\subsection{General Early Exit (gNet) Implementation}
We create the model according to the description in section \ref{gnet_} and train it according to section \ref{training}. 
We use  a stochastic gradient descent optimizer, Adam optimizer, with a default learning rate of $5\times10^{-3}$.
While training, we monitored the validation loss (depending on the application and pre-trained network) to avoid overfitting. We saved the model if the validation loss decreased, and if the validation loss kept increasing for $10$ consecutive epochs, we stopped the training. The best-saved model was used for the inference. To be consistent with prior works~\cite{teerapittayanon2016branchynet, montanari2020eperceptive}, we add exit points after the convolutional layers only. However, our proposed \exitname supports optimal exit points described in ~\cite{jeon2023harvnet}.

\subsection{\compname Implementation}
\compname is implemented as a plug-and-play tool that takes a Pytorch model and the different datasets needed for training (training and test sets). A validation set can optionally be used to evaluate the model during compression and identify the best one. 
In the case of overfitting, it is possible to specify a regularization term, which will be used in the backpropagation phases.
\compname automatically excludes norm layers or biases from compression since they have a low number of parameters and can have huge impacts on accuracy. For gradient and backpropagation operations, \compname uses PyTorch’s automatic differentiation engine Autograd, which allows compressing and retraining of any Pytorch model. 
In the case of models that require data preprocessing for the inferences or that use unconventional forward functions, it is also possible to inject both functions during the compression phase. The preprocessing function is be applied on each input batch before the forwarding phase, while the forward function will be applied on the input instead of the standard one. These features allow unconventional models, such as \exitname, to be compressed without any changes.

\begin{table*}[!htb]
\centering
\caption{Pre-trained DNNs considered in this section.}
\resizebox{0.8\textwidth}{!}{
\label{tab:dataset}
\begin{tabular}{>{\columncolor{black!3}}c c|  >{\columncolor{black!3}}c c |>{\columncolor{black!3}}c c}
\toprule	
\multicolumn{2}{c|}{\textbf{CIFAR-10}\cite{krizhevsky2009learning}}   & \multicolumn{2}{c|}{\textbf{KWS}\cite{warden2018speech}} & \multicolumn{2}{c}{\textbf{HAR}\cite{har_dataset}}\\ 
\rowcolor{gray!6}
\multicolumn{1}{c}{\textbf{Micro}} & \multicolumn{1}{c|}{\textbf{Tiny}} & \multicolumn{1}{c}{\textbf{Micro}} & \multicolumn{1}{c|}{\textbf{Tiny}} & \multicolumn{1}{c}{\textbf{Micro}} & \multicolumn{1}{c}{\textbf{Tiny}} \\ 
\midrule
 C:$64\times3\times3\times3$         & C:$32\times3\times3\times3$           & C:$16\times1\times3\times3$       & C:$16\times1\times3\times3$           & C:$32\times3\times1\times12$           & C:$32\times3\times1\times12$         \\
 C:$128\times64\times3\times3$       & C:$32\times32\times3\times3$          & C:$32\times16\times3\times3$      & C:$32\times16\times3\times3$          & C:$32\times32\times1\times12$          & C:$32\times32\times1\times12$        \\
 C:$64\times128\times3\times3$       & C:$64\times32\times3\times3$          & C:$64\times32\times3\times3$      & C:$64\times32\times3\times3$          & C:$64\times32\times1\times12$         & C:$64\times32\times1\times12$        \\
 F:$256\times256$                    & C:$64\times64\times3\times3$          & F:$2304\times64$                  & C:$64\times64\times3\times3$          & F:$1600\times128$                    & C: $64\times64\times1\times3$        \\
 F:$256\times64$                     & C:$128\times64\times3\times3$         & F:$64\times10$                    & F:$256\times128$                      & F:$128\times128$                       & C:$128\times64\times1\times12$       \\
 F:$64\times10$                      & C:$128\times128\times3\times3$        &                                   & F:$128\times64$                       & F:$128\times6$                        & C:$128\times128\times1\times12$      \\
                                    & F:$2048\times128$                     &                                   & F:$64\times10$                        &                                     & $4\times$ F:$128\times128$      \\
                                    & $3\times$ F:$128\times128$                      &                                   &                                       &                                     & F:$128\times6$                     \\
                                    & F:$128\times10$                       &   C: Convolution Layer                                &         F: Fully Connected Layer                              &                                     &                        \\ 
\bottomrule
\end{tabular}
}
\vspace{-1.5em}
\end{table*}

\subsection{Model Code Generation}

After compressing and adding early exit layers, \sysname converts the model parameters from Python into C headers per layer. To store the weights of the model we used the CSR (Compressed Sparse Row) representation, which allows computationally
efficient matrix operations. In CSR representation, a matrix is flattened and only the non-zero values are saved along with their column indices and extents of rows. It is also possible to choose different representations
e.g., pair representation (index, value), CSC (Compressed Sparse Column), and COO (Coordinate Format)~\cite{han2015deep}.

\subsection{Intermittent Execution Runtime}

There are recent works that provided basic ML library implementations that can be executed intermittently on MSP430FR~\cite{msp430fr59xxx} series MCUs. Therefore, we did not implement an ML library from scratch and relied on existing publicly-available code targeting intermittent systems. For instance, Sonic~\cite{gobieski2019intelligence} provided a task-based implementation of ML operations based on Alpaca~\cite{maeng2017alpaca}. Similarly, ImmortalThreads~\cite{yildiz2022immortal} has also a checkpoint-based ML library. 
We used and modified the ML library implementation of Sonic for the ease of portability since the ImmortalThreads library had some platform-specific assembly code in its source. The SONIC ML library keeps the model parameters in non-volatile memory by using specific structures. We further process the C header outputs of \compname and make them compatible with the SONIC ML library. We implemented a small runtime on top of the Alpaca for Sonic to execute models layer-by-layer and perform runtime optimizations and decisions for \exitname as mentioned in Section~\ref{sec:runtime}.




\section{Evaluation}
\label{sec:eval}

We consider neural network models of two sizes -- (1) \textbf{tiny} models that are suitable for constrained embedded systems 
, and (2) \textbf{micro} models which can fit in extremely constrained systems. 
Table~\ref{tab:dataset} shows the details of all networks. For training, we use batch normalization and RELU activation after each convolutional layer for all these models. To reduce overfitting, we use dropout between the fully connected layers. We train these networks on a GPU machine with two RTX 3090Ti. These models are considered the \emph{pre-trained} models for our evaluation. 

We aim to test the effectiveness of FreeML on datasets from different domains. Firstly, we conduct all our experiments on an image classification dataset ( CIFAR-10 \cite{krizhevsky2009learning}), an acoustic dataset (Google Keyword Spotting (KWS)\cite{warden2018speech}), and a motion-based human activity recognition dataset (HAR~\cite{har_dataset}. We divide the training dataset into a train set and a test set. After shuffling all the training datasets, we put $90\%$ data into the train set and the remaining $10\%$ in the validation set. The test dataset remains unseen throughout the training sessions and is only used for inference.

\subsection{Evaluation of Compression}

\subsubsection{Performance Metrics.} To evaluate the performance of \compname we report two metrics: accuracy and compression rate. Accuracy is used to compare the performance of a model before and after compression. The compression rate indicates how significant the pruning is, and it is calculated by dividing the original number of parameters by the number of pruned parameters. 

\begin{table}[!htb]
\vspace{-1.2em}
\centering
\scriptsize
\caption{Comparison between Genesis and \compname
}
\begin{tabular}{ >{\centering}m{0.08\textwidth}
>{\centering}m{0.04\textwidth}
>{\centering}m{0.045\textwidth}
>{\centering}m{0.04\textwidth}
>{\centering}m{0.045\textwidth}
>{\centering}m{0.04\textwidth}
m{0.05\textwidth}<{\centering}}
\toprule
\bf Network & \multicolumn{2}{c}{\bf Original Model}  & \multicolumn{2}{c}{\bf Genesis} & \multicolumn{2}{c}{\bf SparseComp} \\
        & Acc.       & Size.    & Acc.   & Size.     & Acc. & Size  \\  
\midrule 
\rowcolor{black!3}
Image Classification (MNIST) & 99.06\% & 1905.3\,kb & 99\% & 34.6\,kb (55$\times$) & 98.6\%         & 19.8\,kb (95.8$\times$)  \\
Human Activity Recognition (HAR) & 91.93\%               &  2100, 6 kb (8.2$\times$)             &  88,0 \%  & 256.8 kb (8.2$\times$)  &  88,55 \%          &   29.8\,kb (70.3$\times$) \\
\rowcolor{black!3}
Google Keyword
Spotting (KWS) & 79.97\% &  1316,5\,kb & 84,0\% & 215.5\,kb (6$\times$) & 75.98\%  & 59.2\,kb (22.2$\times$)\\
\bottomrule
\end{tabular}
\label{tab:genesis_vs_sparsecomp_models}
\vspace{-1.2em}
\end{table}

\begin{figure*}[!htb]
\begin{minipage}{0.3\textwidth}
\centering
    \includegraphics[width=\textwidth]{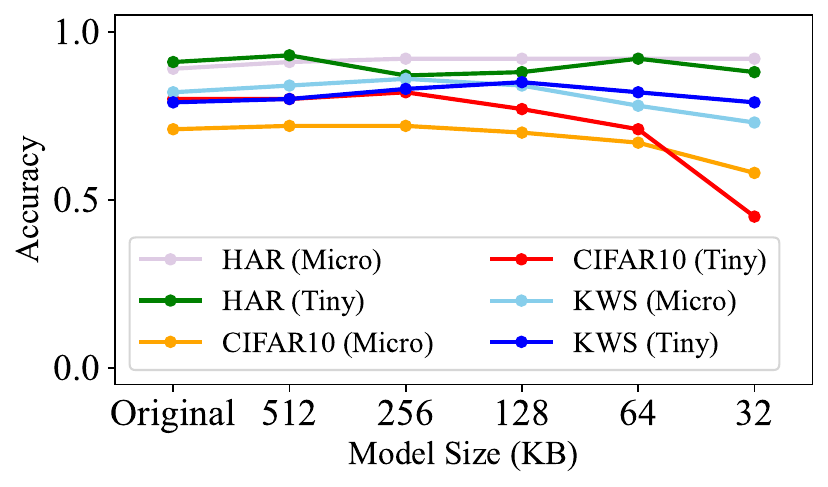}
    \caption{Accuracy of compression. 
    }
    \label{fig:compression_vs_accuracy}
\end{minipage}
\begin{minipage}{0.69\textwidth}
\centering
    \includegraphics[width=\textwidth]{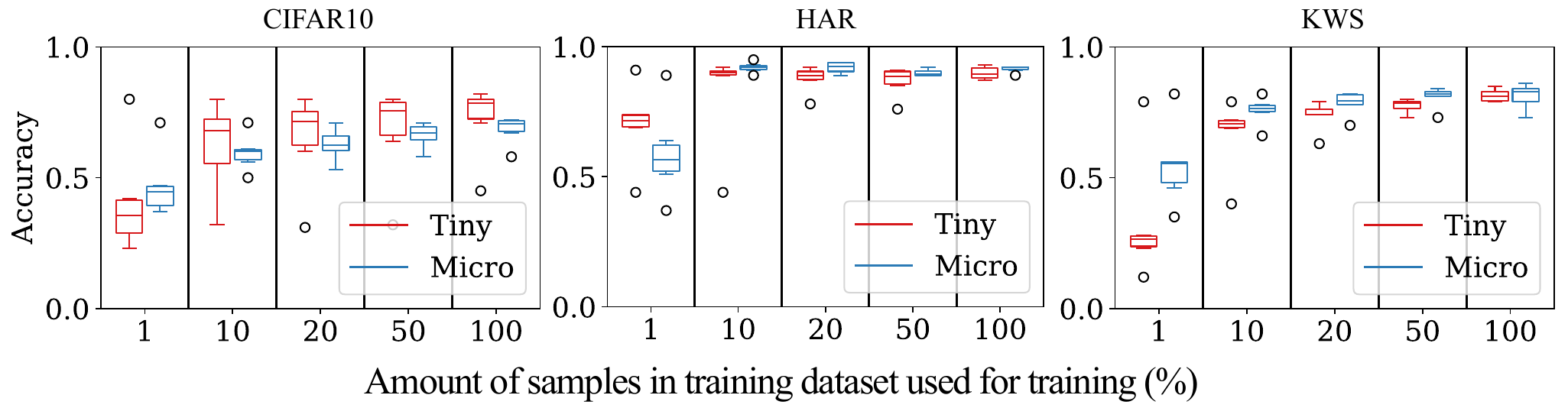}
    \caption{Effect of training dataset percentage on \compname compression.}
    \label{fig:compression_and_dataset_usage}
\end{minipage}
\vspace{-1em}
\end{figure*}

\subsubsection{Comparison against GENESIS}

We used Genesis, the de facto model optimization tool in intermittent computing, as a baseline. \compname uses a simple approach for each model and each layer compared to Genesis, which employs a NAS-oriented approach and employs separation and post-facto pruning. We considered the same datasets and model structures reported by authors of Genesis~\cite[Table 2]{gobieski2019intelligence}. 
Table \ref{tab:genesis_vs_sparsecomp_models} shows the size and accuracy of the uncompressed original models as well as the overall accuracy and size together with the compression rate of the compressed versions. 
On the MNSIT and HAR datasets \compname achieves significantly better compression while maintaining similar precision. 
For KWS, even though we used the same model structure reported in ~\cite[Table 2]{gobieski2019intelligence}, even our uncompressed original model could not reach 84\% accuracy of the compressed model reported by the authors of Genesis. Therefore, for the KWS model, we could not make a sound comparison against Genesis. In our case, when we fix the size of our compressed model to the same size as the compressed model in Genesis, we achieved an accuracy of 79,97,  which is 4\% less than the accuracy reported by Genesis. In this case, the accuracy drop from the uncompressed model was only $~1$ \%. We further proceed with the compression and observed that \compname can compress the original model almost 22 times smaller with under 4\% accuracy drop. Our evaluation showed that \compname is a significantly better and simpler approach compared to existing solutions.

\subsubsection{General Evaluation}
We evaluated \compname using the models shown in Table~\ref{tab:dataset}. 
We would like to highlight that \compname can be applied to compress different models and DNNs without any modification.  Figure~\ref{fig:compression_vs_accuracy} represents the accuracy of the models over several compression iterations. 
At each step, we define a memory size constraint, and \compname compresses the models to meet that requirement.
We observed that for most of the models, \compname obtains promising results with a high compression rate. The accuracy drop is almost linear even reaching kB-sized models. Since \compname performs re-training, at some iterations, we even observed better accuracy with a smaller model.
It is worth mentioning that uncompressed models with higher accuracy lead to higher and more stable accuracy during the compression.


\subsubsection{Effect of Available Training Samples on \compname.} In this section, we evaluate the impact of using only a subset of the dataset on the compression performance. Based on Figure~\ref{fig:compression_and_dataset_usage}, \compname can compress a pre-trained model up to 64 kb without significantly affecting accuracy, even when using only a fraction of the training data. However, when compressing to 32 kb, the performance loss is more noticeable, and a larger pool of training examples is more effective. It's worth noting that the dataset we considered initially had a small number of examples. With larger datasets, the percentage of usage can be even lower.

\begin{figure*}
\centering
    \includegraphics[width=\textwidth]{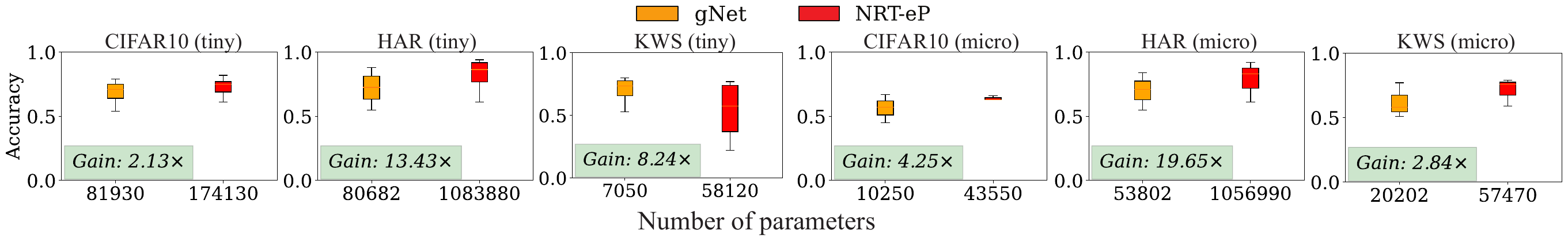}
    \caption{No. of parameters for \exitname and NRT-eP. \exitname requires fewer parameters than NRT-eP, reducing memory overhead.}
    \label{fig:gnet_params}
\vspace{-1em}
\end{figure*}
\begin{figure*}
\centering
    \includegraphics[width=\textwidth]{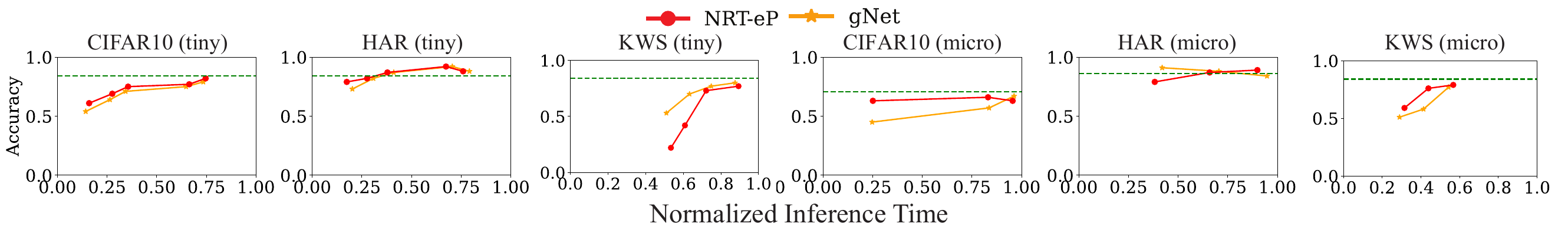}
    \caption{Accuracy and normalized inference time comparison of gNet and NRT-eP with $6$ pre-trained models and $3$ datasets. Here, the dashed green line shows the baseline models' accuracy.}
    \label{fig:gnet_acc}
\vspace{-1em}
\end{figure*}
\begin{figure*}[!htb]
\begin{minipage}{0.23\textwidth}
\centering
    \includegraphics[width=\textwidth]{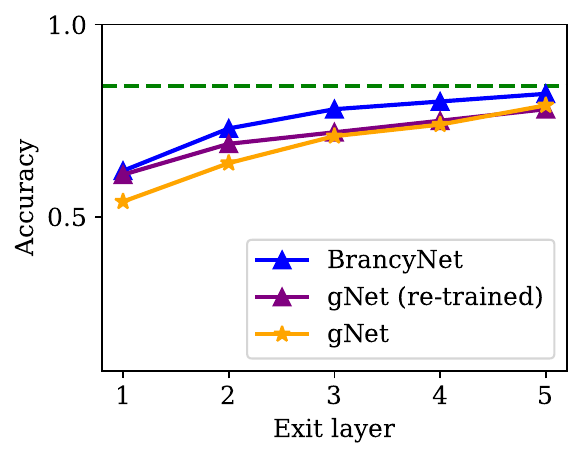}
    \caption{Effect of retraining the baseline on \exitname.}
    \label{brancynet_comparison}
\end{minipage}
\begin{minipage}{0.69\textwidth}
\centering
    \includegraphics[width=\textwidth]{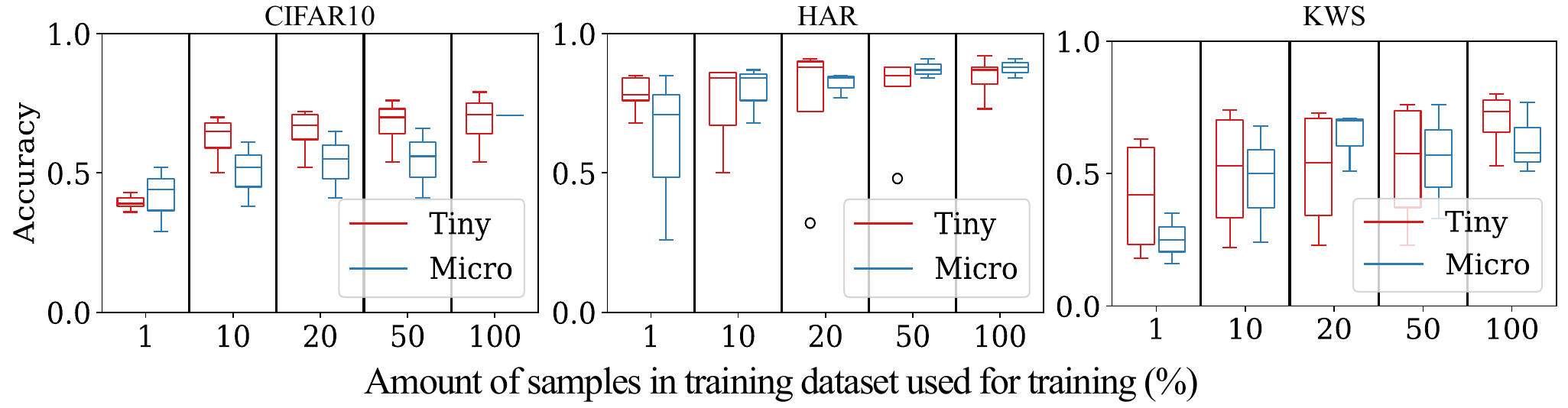}
    \caption{Effect of training dataset percentage on \exitname with CIFAR-10, HAR, and KWS dataset.}
    \label{act_dataset_pct}
\end{minipage}
\vspace{-1.5em}
\end{figure*}

\subsection{Global Early-Exit (\exitname) Evaluation}
This section focuses on evaluating the performance of the global early exit network (\textbf{gNet}). 
We compare gNet with a variant of state-of-the-art e-Perceptive \cite{montanari2020eperceptive}, where we do not retrain the baseline model (compressed with \compname) for a fair comparison and call this Not Re-Trained e-Perceptive (NRT-eP). Finally, we provide an ablation with a re-trained baseline model (BancyNet \cite{teerapittayanon2016branchynet}).



\subsubsection{Performance Metric.}
\label{metrics}


We used memory, throughput, and performance as metrics to evaluate the performance of gNet from two aspects. First, we compare the \textbf{number of required parameters} between the baseline algorithm and \exitname, which translates to the model's memory requirement. 
Next, we measure the time required to infer the model to quantify throughput. Instead of reporting the absolute time, we report the \textbf{normalized inference time} for a more fair and device-agnostic comparison. To calculate the normalized inference time, we normalize the inference times against the inference time of the pre-trained models. In other words, we consider the inference time of the pre-trained models as 1. Smaller normalized inference time indicated high throughput and less runtime overhead. Finally, to assess the performance, we report the model's \textbf{accuracy}.

\subsubsection{Memory and Accuracy Trade-Off.}

Baseline individual early-exit models (NRT-ep) store $n$ number of different exit models for a $n$ layered DNN, which makes it inefficient in terms of memory overhead. On the other hand, \exitname is a single standalone model that works regardless of the number of layers in the baseline model. This makes the \exitname memory efficient and more suitable for small and edge device-level deployment. Figure \ref{fig:gnet_params} shows that \exitname requires $2.13-19.65$ times fewer parameters than NRT-ePs while maintaining the accuracy distributions at different existing layers. The models with the HAR dataset have the highest memory overhead reduction compared to the other datasets due to the smaller input size of baseline HAR models (tiny, micro). As the input shape is smaller, the intermediate feature sizes are smaller, too. Additionally, after passing the augmentation with a zero-padding layer, the feature shapes get a further reduction. This enables us to design a small \exitname model, which results in a very high gain in terms of parameters. This satisfies our claim of one-model-fit for all.

\subsubsection{Inference Time and Accuracy Trade-Off.}

Figure~\ref{fig:gnet_acc} shows the accuracy vs the normalized inference time required for all dataset and pre-trained model configurations described in Table~\ref{tab:dataset}. For all three tiny models, \exitname achieves $90^{th}$ percentile accuracy gain of $0.12 - -0.31\%$ while reducing the inference time by $2.41\%-3.17\%$. However, for micro models, this accuracy gain reduces to $0.12 - -0.31\%$ with a $2.52\%-4.85\%$ reduction in inference time. Due to the larger depth of the tiny baseline models, the global exit branch of tiny models has a larger input size, and thus, the reduction of inference time is lower than in micro models. However, these more comprehensive intermediate features result in higher accuracy gain. On the other hand, micro baseline models are smaller, generating smaller intermediate features as input to the exit branch, and thus, we experience a higher reduction of inference time. In summary, along with reducing the memory overhead by up to 19.65\%, we reduce inference time with negligible to no accuracy loss.




\subsubsection{Comparison with Jointly-Trained BranchyNet.} 
Figure~\ref{brancynet_comparison} compares \exitname with the popular jointly trained early exit method, BranchyNet~\cite{teerapittayanon2016branchynet} on the tiny model with CIFAR-10 dataset. BrancyNet jointly trains the exit branches are trained alongside the baseline model. Though BranchyNet gains accuracy by at least 3.65\% compared to \exitname, it required $4.42$ times fewer parameters than BranchyNet. We have also jointly retrained \exitname, and the accuracy increases by up to $6.5$\% than \exitname.


\subsubsection{Effect of Available Training Samples.} 
This section evaluates the effect of available training data samples required to train \exitname. This analysis is crucial as, in real-world scenarios, the full training dataset may not be available to train \exitname. 
Figure~\ref{act_dataset_pct} shows that the accuracy of all models increases with the availability of more samples in the training set. However, the gain from 10\% available samples and 100\% samples is not that significant. The median accuracy gain from 10\% sample to 100\% sample is only $4.0\%--20.49\%$ for tiny models and {$4.00\%--19.99\%$} for micro models over all datasets. However, for micro models, this average gain is much higher as intermediate features of micro are less comprehensive than tiny models, \exitname can achieve higher accuracy gain.  More complex datasets, e.g., KWS, require more training data to reach higher accuracy, and thus the accuracy gain between 10\% to 100\% training samples are upto $20.49\%$.



\begin{figure}
\includegraphics[width=\columnwidth]{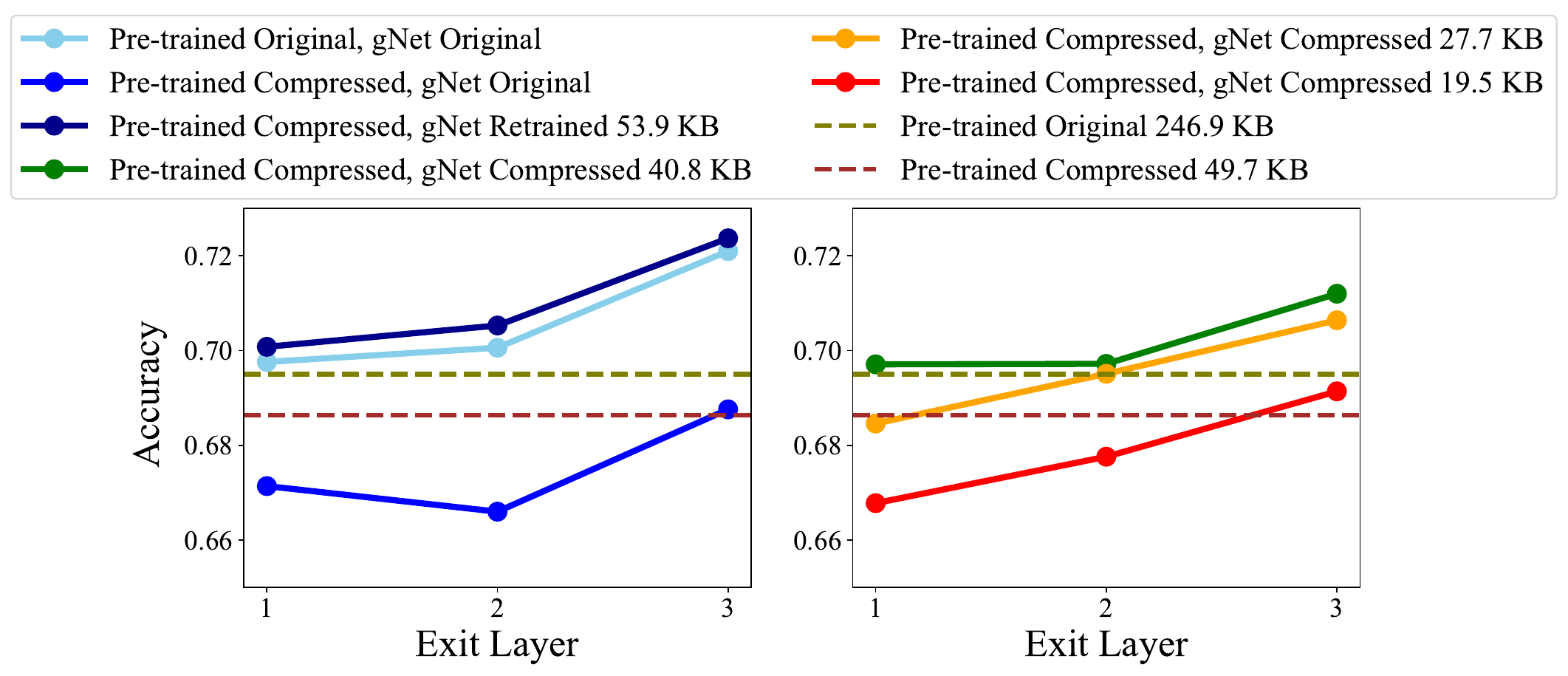}
    \caption{Compression of Early Exit. gNet benefits and performance are maintained during compression.}
    \label{fig:compression_earlyexit}
\vspace{-1.1em}    
\end{figure}

\subsubsection{Evaluation of Early-Exit Compression}
We used CIFAR10 dataset with the aim of compressing models to meet the memory requirements for the MSP430FR device. The compression targets both the baseline model and the gNet with, starting sizes of  $246.9$ KB and $53.9$ KB, respectively. The first step is to compress the baseline model, \compname is able to shrink the sizes to $49.7$ KB with $~1 \%$ accuracy loss. Figure \ref{fig:compression_earlyexit} on the left shows the performance of gNet when relying on the original or the compressed baseline model for the intermediate results. Without a retraining phase, the gNet achieves accuracy comparable to the base model. When it is retrained, however, it even surpasses the performance of the original uncompressed models.
On the right, instead, are shown different compression applied on gNet. The intermediate results are provided by the compressed model and it can be seen that the key features of gNet are maintained during the compression phase by \compname. The two compressed models outperform the base model and reach the accuracy of the uncompressed versions, with a total size reduced from $300.8$ to $77.4$ KB allowing their use on the memory-constrained MSP device.


\begin{table}
	\centering
	\scriptsize

	\caption{Time and energy consumption during the continuous and intermittent execution on MSP430FR5994.}
	\begin{tabular}{ 
             >{\centering}m{0.05\textwidth}
            >{\centering}m{0.04\textwidth}
            >{\centering}m{0.04\textwidth}
            >{\centering}m{0.05\textwidth}
            >{\centering}m{0.04\textwidth}
            >{\centering}m{0.04\textwidth}
            m{0.05\textwidth}<{\centering}}
		\toprule
            \textbf{Network} &\multicolumn{3}{c}{\textbf{Exec. Time (sec)}} & \multicolumn{3}{c}{\textbf{Energy Cons. (mJ)}}  \\
            &\textbf{Cont.} &\textbf{Int.} &\textbf{E.E.} & \textbf{Cont.} &\textbf{Int.} &\textbf{E.E.}  \\
			
		\midrule
             \textbf{Cifar-10} & 139.5 & 166.2 & 156.1 (E3) & 264.9 & 320.1 & 297.3 (E3)\\ 
             \textbf{HAR} & 74.6 & 83.4 & 75.8 (E5) & 135.8 & 152.3 & 137.2 (E5) \\ 
             \textbf{KWS} & 184.6 & 218.7 & 185.8 (E2) & 355.2 & 419.9 & 356.5 (E2)\\ 
    	\bottomrule
	\end{tabular}
	\label{tab:hardware_time_energy_mem}
\vspace{-2em}  
\end{table}

\begin{table}
	\centering
	\scriptsize
	\caption{Total memory overhead of \sysname and tiny models}
	\begin{tabular}{ 
             >{\centering}m{0.06\textwidth}
            >{\centering}m{0.06\textwidth}
            >{\centering}m{0.06\textwidth}
            >{\centering}m{0.06\textwidth}
            m{0.06\textwidth}<{\centering}}
		\toprule
            &\textbf{\sysname} &\multicolumn{3}{c}{\textbf{ Models}} \\
            & & \textbf{Cifar-10} &\textbf{HAR} &\textbf{KWS}  \\
			
		\midrule
             \textbf{.text} & 12.5 kb & 376 Byte & 362 Byte & 334 Byte\\ 
             \textbf{RAM} &4 Byte & 0 & 0 & 0\\ 
             \textbf{FRAM} & 12.8 kb & 200.9 kb & 30.4 kb & 219.4 kb \\ 
    	\bottomrule
	\end{tabular}
	\label{tab:hardware_mem}
\vspace{-3em} 
\end{table}

\subsection{Model Execution on Real Hardware}

We evaluated tiny versions of the gNet models for Cifar-10, HAR, and KWS using MSP430FR5994~\cite{msp430fr59xxx}, the defacto MCU in batteryless systems,  configured to operate at 1 MHz. We report the memory overhead of these models and their total energy consumption and execution time under continuous and intermittent energy supply. For energy harvesting, we used  Powercast TX91501-3W-ID power transmitter and P2110-EVB power harvester that includes a 1mF onboard energy storage supercapacitor.

\begin{figure}
\centering
    \includegraphics[width=0.7\columnwidth]{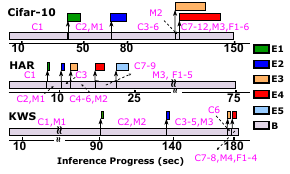}
    \caption{Time overheads of the inference layers on continuous power (C:Convolution, F:Fully Connected, M:Maxpool).}
    \label{fig:infer-progress}
\vspace{-1.2em}    
\end{figure}

\subsubsection{Time and Energy Performance.} 

\label{sec:hw-time}
Table~\ref{tab:hardware_time_energy_mem} shows the execution time and energy consumption of these models on continuous power (``Cont.'' column) as well as under intermittent power {\em without} exit branches (``Int'' column), and {\em with} early exit branches (``E.E.'' column). Our results indicate that early exit layers considerably improve the time and energy overheads of intermittent execution (as evident from the ``Int'' and ``E.E.'' columns of Table~\ref{tab:hardware_time_energy_mem}). For intermittent execution with exit branches, we used the time it takes to run a model continuously as a time constraint to trigger anytime outputs to keep things simple. The idea was to make our models output predictions as fast as their execution on continuous power. When executing HAR and KWS, \sysname runtime used exit branches 5 (E5) and 2 (E2) in these models, respectively, significantly improving their intermittent execution times. For Cifar-10, \sysname used a relatively earlier exit branch 3 (E3), which led to intermittent execution of the early exit branch taking even less time than intermittent execution without using early exits. Figure~\ref{fig:infer-progress} provides a detailed view of the time overheads of each individual layer on continuous power. It shows that the execution times of early exit branches are relatively smaller than those of later exit branches, which is consistent with our arguments in Section~\ref{sec:runtime}.

\subsubsection{Harvested Power and Early Exits.} 
\begin{figure}
   \includegraphics[width=0.9\columnwidth]{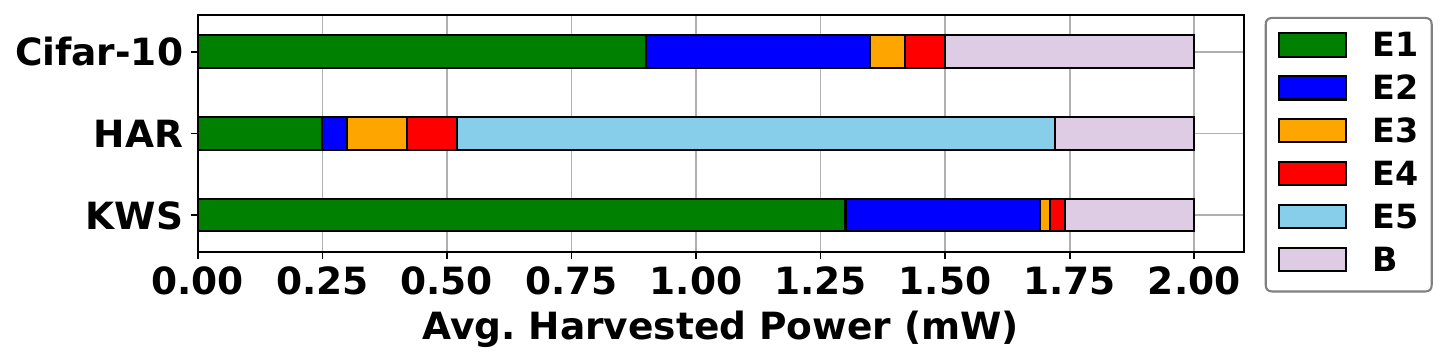}
    \caption{Selected exits based on the average harvested power for the 1mF capacitor of the P2110-EVB RF harvester.}
    \label{fig:power-layers-relation}
\vspace{-1em}    
\end{figure}

To observe how ambient power affects the exit branch taken, we emulated the energy harvesting process of the Powercast receiver to charge a 1mF onboard supercapacitor with different input power levels. 
During our emulations, we imposed the same time constraint as in Section~\ref{sec:hw-time} and also considered the intermittent execution of our models with early exit branches. Figure~\ref{fig:power-layers-relation} presents the exit branches taken with the average input power values ranging from 0 to 2 mW during our emulations. With Cifar-10, \sysname runtime triggers the first exit when the average power input is under approximately 0.85mW to produce results within the time constraint. As the input power increases, model layers are executed faster due to faster charging time, and \sysname runtime takes later exit branches to output predictions with higher accuracy. When the average output power is higher than 1.5mW, the latest exit branch is no longer beneficial since the rest of the model can be executed faster, and the best accuracy can be obtained. Similar behaviors can also be observed for KWS and HAR models.

\subsubsection{Memory Requirements} Table~\ref{tab:hardware_mem} shows the memory consumption of our deployed models on MSP430FR5994 that has a 256KB FRAM and 8KB SRAM. Our compression has reduced the memory consumption of machine learning models so that it can fit in resource-constrained embedded sensing devices. As seen in Table~\ref{tab:hardware_mem}, FRAM usage is within the limit available to the device. FRAM usage depends on the number of model parameters and the layer's input/output size of models. Our compression technique ensures that the layer's size fits in the device.  It must be noted that these numbers also account for the double required to ensure the correct resumption of application under intermittent execution mode and are part of the \sysname library, which is placed in the .text section.


\subsection{Beyond Intermittent Computing}
\begin{figure}
\centering
\includegraphics[width=0.47\textwidth]{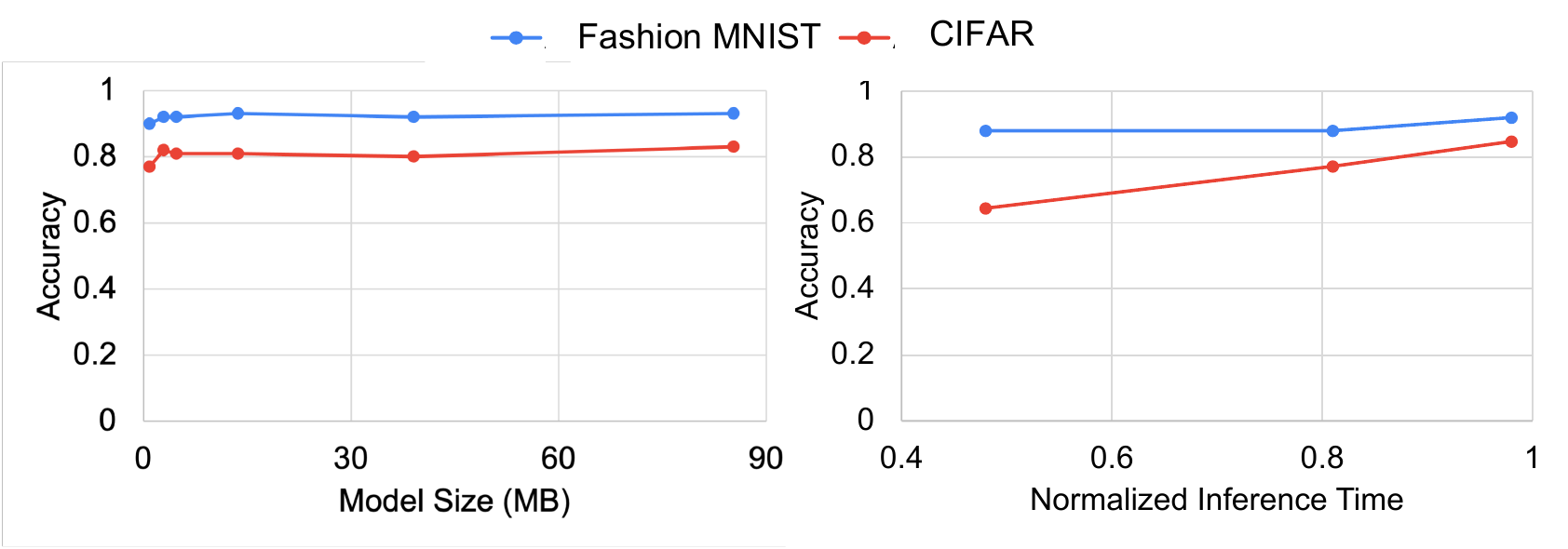}
\caption{Both \compname (left) and \exitname (right) maintains the performance for deeper ResNet34 network.}
\label{fig:beyond}
\vspace{-2em}
\end{figure}

Though we focus on batteryless devices in this work, we envision that the benefit of our proposed \compname and \exitname can go beyond intermittent computing and batteryless devices without any modification. We evaluate these algorithms on deeper ResNet34~\cite{he2015deep} pre-trained models with two datasets (CIFAR-10 and FashionMNIST~\cite{DBLP:journals/corr/abs-1708-07747}) suitable for mobile edge devices to validate our vision. 
We assess the model's effectiveness using the normalized inference time, the number of parameters (described in Section~\ref{metrics}), and accuracy. 
\compname applies pruning to the Residual Block in the same way it does on conventional networks, making it a one-for-all pruning method. Figure~\ref{fig:beyond}(left) shows that \compname compresses the models up to 94 times without a significant drop in accuracy.
Figure~\ref{fig:beyond}(right) demonstrates that \exitname reduces the inference time by $2\%$-$52\%$ and memory-overhead by $7.86-3.93$ times without significantly sacrificing the accuracy.



\section{Discussion and Future Work}
\label{sec:discussion}

\noindpar{Other Platforms.}  We tested deploying compressed models to Apollo 4 Blue Plus which has an ARM-Cortex M4 with a floating point unit and comes with an internal 2 MB MRAM(Magnetic Random-Access Memory) thus providing an opportunity to speed up complex computations. We were able to successfully deploy our model on Apollo 4 using Ambiq's NeuralSPOT~\cite{neuralspot} AI enablement library. ML model deployment using a software development kit prioritizes ease of use over runtime cost and memory footprint optimization, resulting in computational overhead. Even with the bare metal version, software support is required to save and restore volatile data onto MRAM at runtime to ensure correct resumption after power failure as, unlike MSP430, read and write variables are not memory mapped and require explicit APIs for read/write access; thus adding significant overhead to the execution time. In the future, we plan to explore this aspect more to show an end-to-end evaluation of our technique on the Apollo4 platform.

\noindpar{Support for More Complex Models.}
Comparatively hard datasets, e.g., CIFAR-100 require deeper and more complex networks (including residual edge or recurrent layers) making them less suitable for intermittent devices. \compname can compress these networks but when they reach the desired memory footprint, the accuracy drop is significant due to the high number of pruned parameters. We only used unstructured pruning to avoid changing the model architectures, even though structured pruning is required to reduce their runtime buffer requirements. In future work, we plan to explore the performance of \compname in structured pruning as well. 




\noindpar{Retraining the Basline Model during Compression.} 
While \compname works on pre-trained models, it is also possible to retrain all model layers at each compression iteration. We can also train a model from scratch during the compression process. 
This can be very effective for simple datasets, but for more complex datasets, e.g., KWS, we found out that it is better to use pre-trained models as it training converges with higher compression rates.

\noindpar{Reducing Input Size to \exitname.} 
One of the shortcomings of \exitname is the comparatively large input to the global exit branch. Though we reduce the MAC operation and even memory overhead during our MCU implementation by ignoring multiplication with zeros, the input size to \exitname is larger for the deeper layers. In the future, we plan to implement attention-based max-pooling which can successfully pool only the informative components from the output of each layer making the input to the exit-branch even smaller. Moreover, we will be able to better differentiate between the `real' and `zero-padded' components in the earlier layers, resulting in a higher accuracy for the earlier layers. This paper focuses on fine-grained early-exit with exit points after each convolution layer, but our proposed \exitname can be extended to coarse-grain early-exit with exit points after selected layers~\cite{jeon2023harvnet}. Besides, we can further add exit points after fully connected layers if the execution time of inferring the fully connected layers are higher than the exit layer.  




\section{Conclusion}
\label{sec:conclusion}
We proposed \sysname, a novel framework that complements a pre-trained ML model with early exits to react to the power availability compresses the model to fit into a modest memory footprint and generates C code for inference to run on a resource-constrained embedded sensing devices sensor node. Our evaluation shows that \sysname can achieve a 95$\times$ compression rate on pre-trained DNN models reducing their size from MBs to KBs so they can easily be deployed on embedded sensing platforms, as demonstrated by our deployment on MSP430FR5994. Furthermore, \sysname achieves memory overhead 59$\times$ while requiring 65\% lesser time for inference while bearing a minimal drop in accuracy. We release \sysname as an open-source framework via~\cite{hidden-repository} for researchers and the widespread adoption of batteryless TinyML.


\section*{Acknowledgments}

\begin{tabular}{l|r}
 \includegraphics[width=0.17\textwidth]{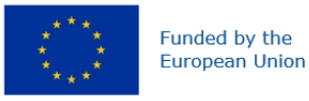} & \includegraphics[width=0.17\textwidth]{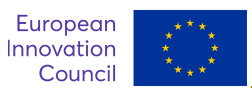}
 \end{tabular} 

\noindent This paper is funded by the European Union (project no. 101071179). Views and opinions expressed are however those of the author(s) only and do not necessarily reflect those of the European Union or EISMEA. Neither the European Union nor the granting authority can be held responsible for them. This paper was also supported, in part, by NSF grants CNS-2347692.

\balance
\bibliography{main}


\begin{thebibliography}{43}


\ifx \showCODEN    \undefined \def \showCODEN     #1{\unskip}     \fi
\ifx \showDOI      \undefined \def \showDOI       #1{#1}\fi
\ifx \showISBNx    \undefined \def \showISBNx     #1{\unskip}     \fi
\ifx \showISBNxiii \undefined \def \showISBNxiii  #1{\unskip}     \fi
\ifx \showISSN     \undefined \def \showISSN      #1{\unskip}     \fi
\ifx \showLCCN     \undefined \def \showLCCN      #1{\unskip}     \fi
\ifx \shownote     \undefined \def \shownote      #1{#1}          \fi
\ifx \showarticletitle \undefined \def \showarticletitle #1{#1}   \fi
\ifx \showURL      \undefined \def \showURL       {\relax}        \fi
\providecommand\bibfield[2]{#2}
\providecommand\bibinfo[2]{#2}
\providecommand\natexlab[1]{#1}
\providecommand\showeprint[2][]{arXiv:#2}

\bibitem[Ahmed et~al\mbox{.}(2019)]%
        {ahmed2019efficient}
\bibfield{author}{\bibinfo{person}{Saad Ahmed}, \bibinfo{person}{Naveed~Anwar
  Bhatti}, \bibinfo{person}{Muhammad~Hamad Alizai},
  \bibinfo{person}{Junaid~Haroon Siddiqui}, {and} \bibinfo{person}{Luca
  Mottola}.} \bibinfo{year}{2019}\natexlab{}.
\newblock \showarticletitle{Efficient intermittent computing with differential
  checkpointing}. In \bibinfo{booktitle}{\emph{Proceedings of the 20th ACM
  SIGPLAN/SIGBED International Conference on Languages, Compilers, and Tools
  for Embedded Systems}}. \bibinfo{pages}{70--81}.
\newblock


\bibitem[Ahmed et~al\mbox{.}(2024)]%
        {ahmed2024internet}
\bibfield{author}{\bibinfo{person}{Saad Ahmed}, \bibinfo{person}{Bashima
  Islam}, \bibinfo{person}{Kasim~Sinan Yildirim}, \bibinfo{person}{Marco
  Zimmerling}, \bibinfo{person}{Przemys{\l}aw Pawe{\l}czak},
  \bibinfo{person}{Muhammad~Hamad Alizai}, \bibinfo{person}{Brandon Lucia},
  \bibinfo{person}{Luca Mottola}, \bibinfo{person}{Jacob Sorber}, {and}
  \bibinfo{person}{Josiah Hester}.} \bibinfo{year}{2024}\natexlab{}.
\newblock \showarticletitle{The Internet of Batteryless Things}.
\newblock \bibinfo{journal}{\emph{Commun. ACM}} \bibinfo{volume}{67},
  \bibinfo{number}{3} (\bibinfo{year}{2024}), \bibinfo{pages}{64--73}.
\newblock


\bibitem[Akhunov and Yildirim(2022)]%
        {akhunov2022adamica}
\bibfield{author}{\bibinfo{person}{Khakim Akhunov} {and}
  \bibinfo{person}{Kasim~Sinan Yildirim}.} \bibinfo{year}{2022}\natexlab{}.
\newblock \showarticletitle{AdaMICA: Adaptive Multicore Intermittent
  Computing}.
\newblock \bibinfo{journal}{\emph{Proceedings of the ACM on Interactive,
  Mobile, Wearable and Ubiquitous Technologies}} \bibinfo{volume}{6},
  \bibinfo{number}{3} (\bibinfo{year}{2022}), \bibinfo{pages}{1--30}.
\newblock


\bibitem[Ambiq(2022)]%
        {neuralspot}
\bibfield{author}{\bibinfo{person}{Ambiq}.} \bibinfo{year}{2022}\natexlab{}.
\newblock \bibinfo{title}{{Ambiq Accelerates the Development of Optimized AI
  Features with neuralSPOT}}.
\newblock
  \bibinfo{howpublished}{\url{https://ambiq.com/news/ambiq-accelerates-the-development-of-optimized-ai-features-with-neuralspot/
  }}.
\newblock


\bibitem[Anonymous(2023)]%
        {hidden-repository}
\bibfield{author}{\bibinfo{person}{Anonymous}.}
  \bibinfo{year}{2023}\natexlab{}.
\newblock \bibinfo{title}{{FreeML Github Repo}}.
\newblock \bibinfo{howpublished}{\url{https://}}.
\newblock


\bibitem[Bakar et~al\mbox{.}(2022a)]%
        {bakar2022protean}
\bibfield{author}{\bibinfo{person}{Abu Bakar}, \bibinfo{person}{Rishabh Goel},
  \bibinfo{person}{Jasper de Winkel}, \bibinfo{person}{Jason Huang},
  \bibinfo{person}{Saad Ahmed}, \bibinfo{person}{Bashima Islam},
  \bibinfo{person}{Przemys{\l}aw Pawe{\l}czak}, \bibinfo{person}{Kas{\i}m~Sinan
  Y{\i}ld{\i}r{\i}m}, {and} \bibinfo{person}{Josiah Hester}.}
  \bibinfo{year}{2022}\natexlab{a}.
\newblock \showarticletitle{Protean: An energy-efficient and heterogeneous
  platform for adaptive and hardware-accelerated battery-free computing}. In
  \bibinfo{booktitle}{\emph{Proceedings of the 20th ACM Conference on Embedded
  Networked Sensor Systems}}. \bibinfo{pages}{207--221}.
\newblock


\bibitem[Bakar et~al\mbox{.}(2022b)]%
        {bakar2022adaptive}
\bibfield{author}{\bibinfo{person}{Abu Bakar}, \bibinfo{person}{Tousif Rahman},
  \bibinfo{person}{Rishad Shafik}, \bibinfo{person}{Fahim Kawsar}, {and}
  \bibinfo{person}{Alessandro Montanari}.} \bibinfo{year}{2022}\natexlab{b}.
\newblock \showarticletitle{Adaptive Intelligence for Batteryless Sensors Using
  Software-Accelerated Tsetlin Machines}. In
  \bibinfo{booktitle}{\emph{Proceedings of SenSys}}.
\newblock


\bibitem[Bakar et~al\mbox{.}(2021)]%
        {bakar2021rehash}
\bibfield{author}{\bibinfo{person}{Abu Bakar}, \bibinfo{person}{Alexander~G
  Ross}, \bibinfo{person}{Kasim~Sinan Yildirim}, {and} \bibinfo{person}{Josiah
  Hester}.} \bibinfo{year}{2021}\natexlab{}.
\newblock \showarticletitle{REHASH: A Flexible, Developer Focused, Heuristic
  Adaptation Platform for Intermittently Powered Computing}.
\newblock \bibinfo{journal}{\emph{Proceedings of the ACM on Interactive,
  Mobile, Wearable and Ubiquitous Technologies}} \bibinfo{volume}{5},
  \bibinfo{number}{3} (\bibinfo{year}{2021}), \bibinfo{pages}{1--42}.
\newblock


\bibitem[Bhattacharya and Lane(2016)]%
        {bhattacharya2016sparsification}
\bibfield{author}{\bibinfo{person}{Sourav Bhattacharya} {and}
  \bibinfo{person}{Nicholas~D Lane}.} \bibinfo{year}{2016}\natexlab{}.
\newblock \showarticletitle{Sparsification and separation of deep learning
  layers for constrained resource inference on wearables}. In
  \bibinfo{booktitle}{\emph{Proceedings of the 14th ACM Conference on Embedded
  Network Sensor Systems}}.
\newblock


\bibitem[Cai et~al\mbox{.}(2022)]%
        {cai2022enable}
\bibfield{author}{\bibinfo{person}{Han Cai}, \bibinfo{person}{Ji Lin},
  \bibinfo{person}{Yujun Lin}, \bibinfo{person}{Zhijian Liu},
  \bibinfo{person}{Haotian Tang}, \bibinfo{person}{Hanrui Wang},
  \bibinfo{person}{Ligeng Zhu}, {and} \bibinfo{person}{Song Han}.}
  \bibinfo{year}{2022}\natexlab{}.
\newblock \showarticletitle{Enable deep learning on mobile devices: Methods,
  systems, and applications}.
\newblock \bibinfo{journal}{\emph{ACM Transactions on Design Automation of
  Electronic Systems (TODAES)}} \bibinfo{volume}{27}, \bibinfo{number}{3}
  (\bibinfo{year}{2022}), \bibinfo{pages}{1--50}.
\newblock


\bibitem[Colin and Lucia(2016)]%
        {colin2016chain}
\bibfield{author}{\bibinfo{person}{Alexei Colin} {and} \bibinfo{person}{Brandon
  Lucia}.} \bibinfo{year}{2016}\natexlab{}.
\newblock \showarticletitle{Chain: tasks and channels for reliable intermittent
  programs}. In \bibinfo{booktitle}{\emph{Proceedings of the ACM SIGPLAN
  International Conference on Object-Oriented Programming, Systems, Languages,
  and Applications}}.
\newblock


\bibitem[Desai et~al\mbox{.}(2022)]%
        {desai2022camaroptera}
\bibfield{author}{\bibinfo{person}{Harsh Desai}, \bibinfo{person}{Matteo
  Nardello}, \bibinfo{person}{Davide Brunelli}, {and} \bibinfo{person}{Brandon
  Lucia}.} \bibinfo{year}{2022}\natexlab{}.
\newblock \showarticletitle{Camaroptera: A long-range image sensor with local
  inference for remote sensing applications}.
\newblock \bibinfo{journal}{\emph{ACM Transactions on Embedded Computing
  Systems (TECS)}} (\bibinfo{year}{2022}).
\newblock


\bibitem[Devices(2020)]%
        {max78000}
\bibfield{author}{\bibinfo{person}{Analog Devices}.}
  \bibinfo{year}{2020}\natexlab{}.
\newblock \bibinfo{title}{Artificial Intelligence Microcontroller with
  Ultra-Low-Power Convolutional Neural Network Accelerator}.
\newblock
\newblock
\urldef\tempurl%
\url{https://datasheets.maximintegrated.com/en/ds/MAX78000.pdf}
\showURL{%
Retrieved June 15, 2022 from \tempurl}


\bibitem[Gobieski et~al\mbox{.}(2019)]%
        {gobieski2019intelligence}
\bibfield{author}{\bibinfo{person}{Graham Gobieski}, \bibinfo{person}{Brandon
  Lucia}, {and} \bibinfo{person}{Nathan Beckmann}.}
  \bibinfo{year}{2019}\natexlab{}.
\newblock \showarticletitle{Intelligence beyond the edge: Inference on
  intermittent embedded systems}. In \bibinfo{booktitle}{\emph{Proceedings of
  the Twenty-Fourth International Conference on Architectural Support for
  Programming Languages and Operating Systems}}. \bibinfo{pages}{199--213}.
\newblock


\bibitem[Gupta et~al\mbox{.}(2017)]%
        {gupta2017protonn}
\bibfield{author}{\bibinfo{person}{Chirag Gupta}, \bibinfo{person}{Arun~Sai
  Suggala}, \bibinfo{person}{Ankit Goyal}, \bibinfo{person}{Harsha~Vardhan
  Simhadri}, \bibinfo{person}{Bhargavi Paranjape}, \bibinfo{person}{Ashish
  Kumar}, \bibinfo{person}{Saurabh Goyal}, \bibinfo{person}{Raghavendra Udupa},
  \bibinfo{person}{Manik Varma}, {and} \bibinfo{person}{Prateek Jain}.}
  \bibinfo{year}{2017}\natexlab{}.
\newblock \showarticletitle{Protonn: Compressed and accurate knn for
  resource-scarce devices}. In \bibinfo{booktitle}{\emph{International
  conference on machine learning}}. PMLR.
\newblock


\bibitem[Han et~al\mbox{.}(2015a)]%
        {han2015deep}
\bibfield{author}{\bibinfo{person}{Song Han}, \bibinfo{person}{Huizi Mao},
  {and} \bibinfo{person}{William~J Dally}.} \bibinfo{year}{2015}\natexlab{a}.
\newblock \showarticletitle{Deep compression: Compressing deep neural networks
  with pruning, trained quantization and huffman coding}.
\newblock \bibinfo{journal}{\emph{arXiv preprint arXiv:1510.00149}}
  (\bibinfo{year}{2015}).
\newblock


\bibitem[Han et~al\mbox{.}(2015b)]%
        {han2015learning}
\bibfield{author}{\bibinfo{person}{Song Han}, \bibinfo{person}{Jeff Pool},
  \bibinfo{person}{John Tran}, {and} \bibinfo{person}{William Dally}.}
  \bibinfo{year}{2015}\natexlab{b}.
\newblock \showarticletitle{Learning both weights and connections for efficient
  neural network}.
\newblock \bibinfo{journal}{\emph{Advances in neural information processing
  systems}}  \bibinfo{volume}{28} (\bibinfo{year}{2015}).
\newblock


\bibitem[He et~al\mbox{.}(2015)]%
        {he2015deep}
\bibfield{author}{\bibinfo{person}{Kaiming He}, \bibinfo{person}{Xiangyu
  Zhang}, \bibinfo{person}{Shaoqing Ren}, {and} \bibinfo{person}{Jian Sun}.}
  \bibinfo{year}{2015}\natexlab{}.
\newblock \showarticletitle{Deep residual learning for image recognition. arXiv
  2015}.
\newblock \bibinfo{journal}{\emph{arXiv preprint arXiv:1512.03385}}
  \bibinfo{volume}{14} (\bibinfo{year}{2015}).
\newblock


\bibitem[Hymel et~al\mbox{.}(2022)]%
        {hymel2022edge}
\bibfield{author}{\bibinfo{person}{Shawn Hymel}, \bibinfo{person}{Colby
  Banbury}, \bibinfo{person}{Daniel Situnayake}, \bibinfo{person}{Alex Elium},
  \bibinfo{person}{Carl Ward}, \bibinfo{person}{Mat Kelcey},
  \bibinfo{person}{Mathijs Baaijens}, \bibinfo{person}{Mateusz Majchrzycki},
  \bibinfo{person}{Jenny Plunkett}, {and} \bibinfo{person}{other}.}
  \bibinfo{year}{2022}\natexlab{}.
\newblock \showarticletitle{Edge Impulse: An MLOps Platform for Tiny Machine
  Learning}.
\newblock \bibinfo{journal}{\emph{arXiv}} (\bibinfo{year}{2022}).
\newblock


\bibitem[Ignatov(n d)]%
        {har_dataset}
\bibfield{author}{\bibinfo{person}{Andrey Ignatov}.} \bibinfo{year}{[n.
  d.]}\natexlab{}.
\newblock \bibinfo{title}{HAR}.
\newblock
\newblock
\urldef\tempurl%
\url{https://github.com/aiff22/HAR}
\showURL{%
\tempurl}


\bibitem[Inc.(2017)]%
        {msp430fr59xxx}
\bibfield{author}{\bibinfo{person}{Texas~Instruments Inc.}}
  \bibinfo{year}{2017}\natexlab{}.
\newblock \bibinfo{title}{MSP430FR59xx Mixed-Signal Microcontrollers (Rev. F)}.
\newblock
\newblock
\urldef\tempurl%
\url{https://www.ti.com/lit/ds/symlink/msp430fr5969.pdf}
\showURL{%
\tempurl}


\bibitem[Islam and Nirjon(2020)]%
        {zygarde2020islam}
\bibfield{author}{\bibinfo{person}{Bashima Islam} {and}
  \bibinfo{person}{Shahriar Nirjon}.} \bibinfo{year}{2020}\natexlab{}.
\newblock \showarticletitle{Zygarde: Time-Sensitive On-Device Deep Inference
  and Adaptation on Intermittently-Powered Systems}.
\newblock \bibinfo{journal}{\emph{Proc. ACM Interact. Mob. Wearable Ubiquitous
  Technol.}} \bibinfo{volume}{4}, \bibinfo{number}{3}.
\newblock


\bibitem[Jeon et~al\mbox{.}(2023)]%
        {jeon2023harvnet}
\bibfield{author}{\bibinfo{person}{Seunghyeok Jeon}, \bibinfo{person}{Yonghun
  Choi}, \bibinfo{person}{Yeonwoo Cho}, {and} \bibinfo{person}{Hojung Cha}.}
  \bibinfo{year}{2023}\natexlab{}.
\newblock \showarticletitle{HarvNet: Resource-Optimized Operation of Multi-Exit
  Deep Neural Networks on Energy Harvesting Devices}. In
  \bibinfo{booktitle}{\emph{Proceedings of the 21st Annual International
  Conference on Mobile Systems, Applications and Services}}.
  \bibinfo{pages}{42--55}.
\newblock


\bibitem[Kim et~al\mbox{.}(2015)]%
        {kim2015compression}
\bibfield{author}{\bibinfo{person}{Yong-Deok Kim}, \bibinfo{person}{Eunhyeok
  Park}, \bibinfo{person}{Sungjoo Yoo}, \bibinfo{person}{Taelim Choi},
  \bibinfo{person}{Lu Yang}, {and} \bibinfo{person}{Dongjun Shin}.}
  \bibinfo{year}{2015}\natexlab{}.
\newblock \showarticletitle{Compression of deep convolutional neural networks
  for fast and low power mobile applications}.
\newblock \bibinfo{journal}{\emph{arXiv preprint arXiv:1511.06530}}
  (\bibinfo{year}{2015}).
\newblock


\bibitem[Krizhevsky(2009)]%
        {krizhevsky2009learning}
\bibfield{author}{\bibinfo{person}{Alex Krizhevsky}.}
  \bibinfo{year}{2009}\natexlab{}.
\newblock \showarticletitle{Learning multiple layers of features from tiny
  images}.
\newblock  (\bibinfo{year}{2009}).
\newblock


\bibitem[Lee and Nirjon(2019)]%
        {lee2019neuro}
\bibfield{author}{\bibinfo{person}{Seulki Lee} {and} \bibinfo{person}{Shahriar
  Nirjon}.} \bibinfo{year}{2019}\natexlab{}.
\newblock \showarticletitle{Neuro. zero: a zero-energy neural network
  accelerator for embedded sensing and inference systems}. In
  \bibinfo{booktitle}{\emph{Proceedings of the 17th Conference on Embedded
  Networked Sensor Systems}}. \bibinfo{pages}{138--152}.
\newblock


\bibitem[Lee and Nirjon(2020)]%
        {lee2020fast}
\bibfield{author}{\bibinfo{person}{Seulki Lee} {and} \bibinfo{person}{Shahriar
  Nirjon}.} \bibinfo{year}{2020}\natexlab{}.
\newblock \showarticletitle{Fast and scalable in-memory deep multitask learning
  via neural weight virtualization}. In \bibinfo{booktitle}{\emph{Proceedings
  of the 18th International Conference on Mobile Systems, Applications, and
  Services}}. \bibinfo{pages}{175--190}.
\newblock


\bibitem[Lee and Nirjon(2022)]%
        {lee2022weight}
\bibfield{author}{\bibinfo{person}{Seulki Lee} {and} \bibinfo{person}{Shahriar
  Nirjon}.} \bibinfo{year}{2022}\natexlab{}.
\newblock \showarticletitle{Weight Separation for Memory-Efficient and Accurate
  Deep Multitask Learning}. In \bibinfo{booktitle}{\emph{2022 IEEE
  International Conference on Pervasive Computing and Communications
  (PerCom)}}. IEEE, \bibinfo{pages}{13--22}.
\newblock


\bibitem[Lin et~al\mbox{.}(2020)]%
        {lin2020mcunet}
\bibfield{author}{\bibinfo{person}{Ji Lin}, \bibinfo{person}{Wei-Ming Chen},
  \bibinfo{person}{Yujun Lin}, \bibinfo{person}{Chuang Gan},
  \bibinfo{person}{Song Han}, {et~al\mbox{.}}} \bibinfo{year}{2020}\natexlab{}.
\newblock \showarticletitle{Mcunet: Tiny deep learning on iot devices}.
\newblock \bibinfo{journal}{\emph{Advances in Neural Information Processing
  Systems}}  \bibinfo{volume}{33} (\bibinfo{year}{2020}),
  \bibinfo{pages}{11711--11722}.
\newblock


\bibitem[Liu et~al\mbox{.}(2018)]%
        {liu2018rethinking}
\bibfield{author}{\bibinfo{person}{Zhuang Liu}, \bibinfo{person}{Mingjie Sun},
  \bibinfo{person}{Tinghui Zhou}, \bibinfo{person}{Gao Huang}, {and}
  \bibinfo{person}{Trevor Darrell}.} \bibinfo{year}{2018}\natexlab{}.
\newblock \showarticletitle{Rethinking the value of network pruning}.
\newblock \bibinfo{journal}{\emph{arXiv preprint arXiv:1810.05270}}
  (\bibinfo{year}{2018}).
\newblock


\bibitem[Maeng et~al\mbox{.}(2017)]%
        {maeng2017alpaca}
\bibfield{author}{\bibinfo{person}{Kiwan Maeng}, \bibinfo{person}{Alexei
  Colin}, {and} \bibinfo{person}{Brandon Lucia}.}
  \bibinfo{year}{2017}\natexlab{}.
\newblock \showarticletitle{Alpaca: Intermittent execution without
  checkpoints}.
\newblock \bibinfo{journal}{\emph{Proceedings of the ACM on Programming
  Languages}} \bibinfo{volume}{1}, \bibinfo{number}{OOPSLA}
  (\bibinfo{year}{2017}), \bibinfo{pages}{1--30}.
\newblock


\bibitem[Majid et~al\mbox{.}(2020)]%
        {majid2020dynamic}
\bibfield{author}{\bibinfo{person}{Amjad~Yousef Majid},
  \bibinfo{person}{Carlo~Delle Donne}, \bibinfo{person}{Kiwan Maeng},
  \bibinfo{person}{Alexei Colin}, \bibinfo{person}{Kasim~Sinan Yildirim},
  \bibinfo{person}{Brandon Lucia}, {and} \bibinfo{person}{Przemys{\l}aw
  Pawe{\l}czak}.} \bibinfo{year}{2020}\natexlab{}.
\newblock \showarticletitle{Dynamic task-based intermittent execution for
  energy-harvesting devices}.
\newblock \bibinfo{journal}{\emph{ACM Transactions on Sensor Networks (TOSN)}}
  \bibinfo{volume}{16}, \bibinfo{number}{1} (\bibinfo{year}{2020}),
  \bibinfo{pages}{1--24}.
\newblock


\bibitem[Mendis et~al\mbox{.}(2021)]%
        {mendis2021intermittent}
\bibfield{author}{\bibinfo{person}{Hashan~Roshantha Mendis},
  \bibinfo{person}{Chih-Kai Kang}, {and} \bibinfo{person}{Pi-cheng Hsiu}.}
  \bibinfo{year}{2021}\natexlab{}.
\newblock \showarticletitle{Intermittent-aware neural architecture search}.
\newblock \bibinfo{journal}{\emph{ACM Transactions on Embedded Computing
  Systems (TECS)}} \bibinfo{volume}{20}, \bibinfo{number}{5s}
  (\bibinfo{year}{2021}), \bibinfo{pages}{1--27}.
\newblock


\bibitem[Montanari et~al\mbox{.}(2020)]%
        {montanari2020eperceptive}
\bibfield{author}{\bibinfo{person}{Alessandro Montanari},
  \bibinfo{person}{Manuja Sharma}, \bibinfo{person}{Dainius Jenkus},
  \bibinfo{person}{Mohammed Alloulah}, \bibinfo{person}{Lorena Qendro}, {and}
  \bibinfo{person}{Fahim Kawsar}.} \bibinfo{year}{2020}\natexlab{}.
\newblock \showarticletitle{ePerceptive: energy reactive embedded intelligence
  for batteryless sensors}. In \bibinfo{booktitle}{\emph{Proceedings of the
  18th Conference on Embedded Networked Sensor Systems}}.
  \bibinfo{pages}{382--394}.
\newblock


\bibitem[Nakajima et~al\mbox{.}(2011)]%
        {nakajima2011global}
\bibfield{author}{\bibinfo{person}{Shinichi Nakajima}, \bibinfo{person}{Masashi
  Sugiyama}, {and} \bibinfo{person}{S Babacan}.}
  \bibinfo{year}{2011}\natexlab{}.
\newblock \showarticletitle{Global solution of fully-observed variational
  Bayesian matrix factorization is column-wise independent}.
\newblock \bibinfo{journal}{\emph{Advances in Neural Information Processing
  Systems}}  \bibinfo{volume}{24} (\bibinfo{year}{2011}).
\newblock


\bibitem[Teerapittayanon et~al\mbox{.}(2016)]%
        {teerapittayanon2016branchynet}
\bibfield{author}{\bibinfo{person}{Surat Teerapittayanon},
  \bibinfo{person}{Bradley McDanel}, {and} \bibinfo{person}{Hsiang-Tsung
  Kung}.} \bibinfo{year}{2016}\natexlab{}.
\newblock \showarticletitle{Branchynet: Fast inference via early exiting from
  deep neural networks}. In \bibinfo{booktitle}{\emph{2016 23rd International
  Conference on Pattern Recognition (ICPR)}}. IEEE,
  \bibinfo{pages}{2464--2469}.
\newblock


\bibitem[Warden(2018)]%
        {warden2018speech}
\bibfield{author}{\bibinfo{person}{Pete Warden}.}
  \bibinfo{year}{2018}\natexlab{}.
\newblock \showarticletitle{Speech commands: A dataset for limited-vocabulary
  speech recognition}.
\newblock \bibinfo{journal}{\emph{arXiv preprint arXiv:1804.03209}}
  (\bibinfo{year}{2018}).
\newblock


\bibitem[Xiao et~al\mbox{.}(2017)]%
        {DBLP:journals/corr/abs-1708-07747}
\bibfield{author}{\bibinfo{person}{Han Xiao}, \bibinfo{person}{Kashif Rasul},
  {and} \bibinfo{person}{Roland Vollgraf}.} \bibinfo{year}{2017}\natexlab{}.
\newblock \showarticletitle{Fashion-MNIST: a Novel Image Dataset for
  Benchmarking Machine Learning Algorithms}.
\newblock \bibinfo{journal}{\emph{CoRR}}  \bibinfo{volume}{abs/1708.07747}
  (\bibinfo{year}{2017}).
\newblock
\showeprint[arXiv]{1708.07747}
\urldef\tempurl%
\url{http://arxiv.org/abs/1708.07747}
\showURL{%
\tempurl}


\bibitem[Yao et~al\mbox{.}(2018)]%
        {yao2018fastdeepiot}
\bibfield{author}{\bibinfo{person}{Shuochao Yao}, \bibinfo{person}{Yiran Zhao},
  \bibinfo{person}{Huajie Shao}, \bibinfo{person}{ShengZhong Liu},
  \bibinfo{person}{Dongxin Liu}, \bibinfo{person}{Lu Su}, {and}
  \bibinfo{person}{Tarek Abdelzaher}.} \bibinfo{year}{2018}\natexlab{}.
\newblock \showarticletitle{Fastdeepiot: Towards understanding and optimizing
  neural network execution time on mobile and embedded devices}. In
  \bibinfo{booktitle}{\emph{Proceedings of the 16th ACM Conference on Embedded
  Networked Sensor Systems}}. \bibinfo{pages}{278--291}.
\newblock


\bibitem[Yao et~al\mbox{.}(2017)]%
        {yao2017deepiot}
\bibfield{author}{\bibinfo{person}{Shuochao Yao}, \bibinfo{person}{Yiran Zhao},
  \bibinfo{person}{Aston Zhang}, \bibinfo{person}{Lu Su}, {and}
  \bibinfo{person}{Tarek Abdelzaher}.} \bibinfo{year}{2017}\natexlab{}.
\newblock \showarticletitle{Deepiot: Compressing deep neural network structures
  for sensing systems with a compressor-critic framework}. In
  \bibinfo{booktitle}{\emph{Proceedings of the 15th ACM Conference on Embedded
  Network Sensor Systems}}. \bibinfo{pages}{1--14}.
\newblock


\bibitem[Y{\i}ld{\i}r{\i}m et~al\mbox{.}(2018)]%
        {yildirim2018ink}
\bibfield{author}{\bibinfo{person}{Kas{\i}m~Sinan Y{\i}ld{\i}r{\i}m},
  \bibinfo{person}{Amjad~Yousef Majid}, \bibinfo{person}{Dimitris Patoukas},
  \bibinfo{person}{Koen Schaper}, \bibinfo{person}{Przemyslaw Pawelczak}, {and}
  \bibinfo{person}{Josiah Hester}.} \bibinfo{year}{2018}\natexlab{}.
\newblock \showarticletitle{Ink: Reactive kernel for tiny batteryless sensors}.
  In \bibinfo{booktitle}{\emph{Proceedings of the 16th ACM Conference on
  Embedded Networked Sensor Systems}}. \bibinfo{pages}{41--53}.
\newblock


\bibitem[Yildiz et~al\mbox{.}(2023)]%
        {yildiz2023efficient}
\bibfield{author}{\bibinfo{person}{Eren Yildiz}, \bibinfo{person}{Saad Ahmed},
  \bibinfo{person}{Bashima Islam}, \bibinfo{person}{Josiah Hester}, {and}
  \bibinfo{person}{Kasim~Sinan Yildirim}.} \bibinfo{year}{2023}\natexlab{}.
\newblock \showarticletitle{Efficient and Safe I/O Operations for Intermittent
  Systems}. In \bibinfo{booktitle}{\emph{Proceedings of the Eighteenth European
  Conference on Computer Systems}}. \bibinfo{pages}{63--78}.
\newblock


\bibitem[Y{\i}ld{\i}z et~al\mbox{.}(2022)]%
        {yildiz2022immortal}
\bibfield{author}{\bibinfo{person}{Eren Y{\i}ld{\i}z}, \bibinfo{person}{Lijun
  Chen}, {and} \bibinfo{person}{Kasim~Sinan Y{\i}ld{\i}r{\i}m}.}
  \bibinfo{year}{2022}\natexlab{}.
\newblock \showarticletitle{Immortal Threads: Multithreaded Event-driven
  Intermittent Computing on $\{$Ultra-Low-Power$\}$ Microcontrollers}. In
  \bibinfo{booktitle}{\emph{16th USENIX Symposium on Operating Systems Design
  and Implementation (OSDI 22)}}. \bibinfo{pages}{339--355}.
\newblock


\end{thebibliography}
\bibliographystyle{ACM-Reference-Format}

\end{document}